\pdfoutput=1

\def\year{2022}\relax
\documentclass[letterpaper]{article} %
\usepackage{aaai22}  %
\usepackage{times}  %
\usepackage{helvet}  %
\usepackage{courier}  %
\usepackage[hyphens]{url}  %
\usepackage{graphicx} %
\usepackage{natbib}  %
\usepackage{caption} %
\DeclareCaptionStyle{ruled}{labelfont=normalfont,labelsep=colon,strut=off} %
\usepackage{algorithm}
\usepackage{algorithmic}

\usepackage[inline]{enumitem}
\usepackage{xcolor}
\usepackage{subcaption}
\usepackage{booktabs}
\usepackage{xspace}
\usepackage{multirow}

\usepackage[activate={true,nocompatibility}, kerning=true,spacing=true, stretch=10,shrink=5]{microtype}
\microtypecontext{spacing=nonfrench}
\usepackage{newfloat}
\usepackage{listings}
\floatstyle{ruled}
\newfloat{listing}{tb}{lst}{}
\floatname{listing}{Listing}

\definecolor{dkgreen}{rgb}{0,0.6,0}

\newenvironment{itemizesquish}{\begin{list}{\setcounter{enumi}{0}\labelitemi}{\setlength{\itemsep}{-0.25em}\setlength{\labelwidth}{0.5em}\setlength{\leftmargin}{\labelwidth}\addtolength{\leftmargin}{\labelsep}}}{\end{list}}

\newcommand*\pct{\scalebox{.8}{\%}}
\newcommand{\pos}[1]{\texttt{#1}}

\newcommand{\legalmove}{LgM\xspace}
\newcommand{\exactmove}{ExM\xspace}
\newcommand{\piecetype}{AP\xspace}

\title{Chess as a Testbed for Language Model State Tracking}

\author{
Shubham Toshniwal\textsuperscript{\rm 1},
Sam Wiseman\textsuperscript{\rm 2},
Karen Livescu\textsuperscript{\rm 1},
Kevin Gimpel\textsuperscript{\rm 1}	
}

 \affiliations{

    \textsuperscript{\rm 1}Toyota Technological Institute at Chicago\\
    \textsuperscript{\rm 2}Duke University\\
    \texttt
    {\{shtoshni, klivescu, kgimpel\}@ttic.edu, swiseman@cs.duke.edu}
}

\begin{document}

\maketitle

\begin{abstract}
Transformer language models have made tremendous strides in natural language understanding tasks. 
However, the complexity of natural language makes it challenging
to ascertain how accurately these models are tracking the world state underlying the text.
Motivated by this issue, we consider the task of language modeling for the game of chess.
Unlike natural language, chess notations describe a simple, constrained, and deterministic domain.
Moreover, we observe that the appropriate choice of chess notation allows for directly probing the world state, without requiring any additional probing-related machinery.
We find that: 
\begin{enumerate*}[label=(\alph*)]
	\item With enough training data, transformer language models can learn to track pieces and predict legal moves with high accuracy when trained solely on move sequences.
	\item %
	For small training sets %
	providing access to board state information during training can yield significant improvements. 
	\item The success of transformer language models is dependent on access to the entire game history i.e.\ ``full attention". Approximating this full attention results in a significant performance drop. 
\end{enumerate*}
	We propose this testbed as a benchmark for future work on the development and analysis of transformer language models.
\end{abstract}

\section{Introduction}

Recently, transformer-based language models 
have stretched notions of what is possible with the simple self-supervised objective of language modeling, becoming
a fixture in state of the art language technologies
\citep{vaswani2017attention, devlin-etal-2019-bert, brown2020language}.
However, the black box nature of these models combined with the complexity of natural language makes it
challenging to measure how accurately they
represent the world state underlying the text.

In order to better measure the extent to which these models can capture the world state underlying the symbolic data they consume, we propose training and studying
transformer language models for the game of chess.
Chess provides a simple, constrained, and deterministic domain where the exact world state is known.
Chess games can also be transcribed exactly and unambiguously using chess notations (Section~\ref{sec:chess}).
Most importantly, the form of chess notations allows us to probe our language models for aspects of the board state using simple prompts (Section~\ref{sec:probing}) and without changing the language modeling objective or introducing any new classifiers.\footnote{Code and data available at - \url{https://github.com/shtoshni/learning-chess-blindfolded}}

Due to the simplicity and precision of chess, we can evaluate language model predictions at a more fine-grained level than merely comparing them to the ground truth.
For example, even if the next move prediction doesn't match the ground truth move, we can still evaluate whether the move is legal given the board state, and if it is illegal, the error can be automatically analyzed (Appendix~\ref{sec:error_analysis}).
Moreover, since world state transitions are deterministic and known, we can  evaluate models using counterfactual queries as well.
Our proposed evaluation sets and metrics are described in Section~\ref{sec:cloze}.

\begin{figure*}
\begin{subfigure}{0.3\textwidth}
   \includegraphics[width=\linewidth]{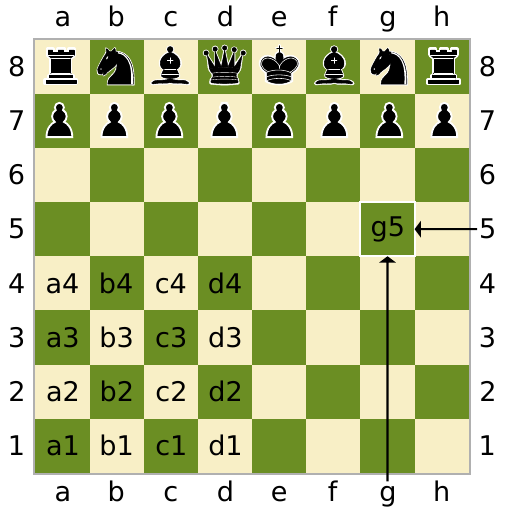}
   \caption{Square naming} \label{fig:algebraic_notation}
\end{subfigure}
\hspace*{\fill}
\begin{subfigure}{0.63\textwidth}
\begin{subfigure}{0.43\textwidth}
   \vspace{.2in}\includegraphics[width=\linewidth]{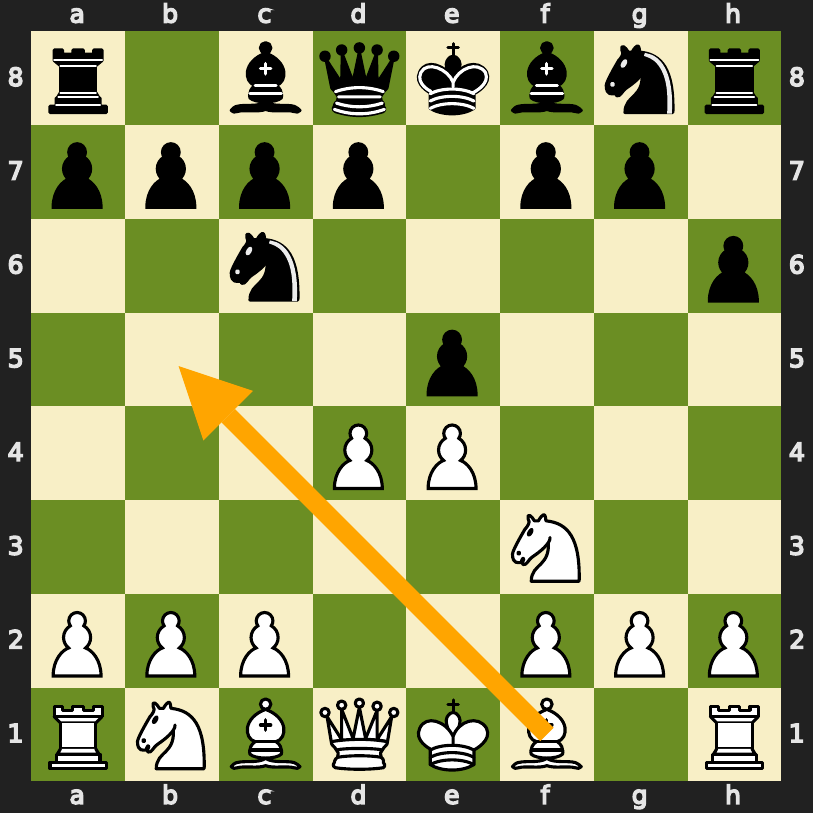}
\end{subfigure}
\hspace*{\fill}
\begin{subfigure}{0.43\textwidth}
   \vspace{.2in}\includegraphics[width=\linewidth]{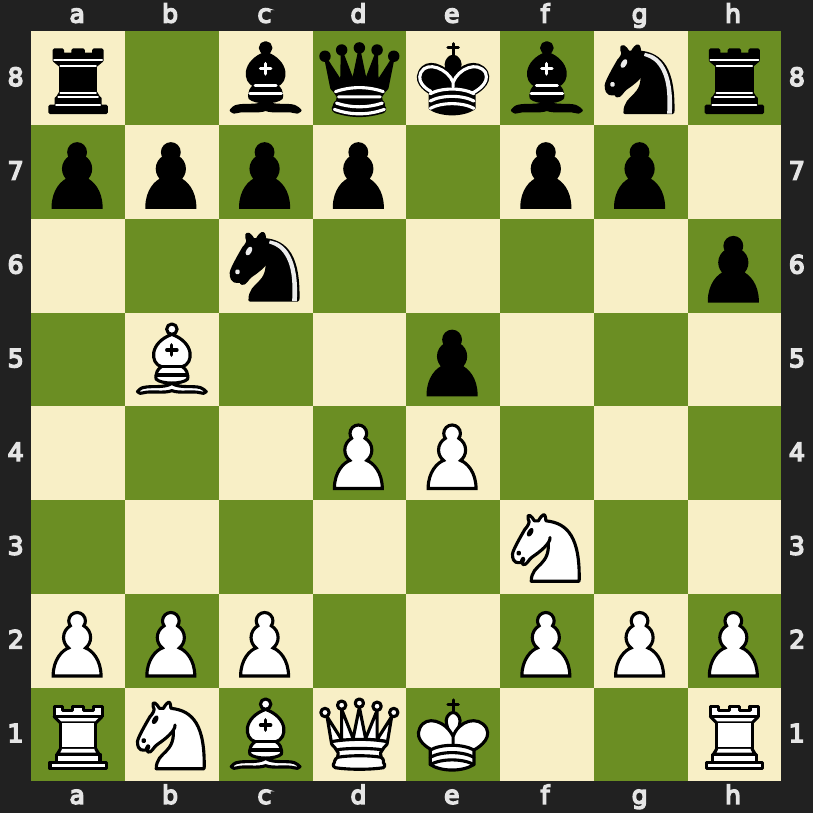}
\end{subfigure}
\vspace{.1in}
\caption{Board state before (left) and after (right) the bishop at \pos{f1} is moved to \pos{b5}. UCI notation represents the move as \pos{f1b5}.}
\label{fig:move_notation}
\end{subfigure}
\vspace{-0.05in}
\caption{
Chess Notation %
}
\end{figure*}

While chess represents a controlled domain,
it is by no means trivial for a language model.
To illustrate the challenges of language modeling for chess,
consider the left board shown in Figure~\ref{fig:move_notation}, where white is next to move.
In order to generate a valid next move, the language model needs to (a) infer that it is white's turn, (b) represent the locations of all pieces, both white and black, (c) select one of the white pieces which can be legally moved, and finally (d) make a legal move with the selected piece.
Thus, a language model has to learn to track the board state, learn to generate moves according to the rules of chess, and on top of that learn chess strategies to predict the actual move. %

We find that when given enough training data, transformers can learn to both track piece locations and predict legal moves with high accuracy.
However, when trained on small training sets, predictive ability suffers. 
In this more challenging setting, introducing parts of the board state as tokens in the training sequences (Section~\ref{sec:rap_board})  improves piece tracking significantly (Appendix~\ref{sec:error_analysis}).

Our results also provide some key insights on transformer language models:
\begin{enumerate*}[label=(\roman*)]
	\item They are robust to changes in input distribution where additional tokens, related to board state, are added to input sequence \emph{only during training} (Section~\ref{sec:rap_board}). 
	In contrast to LSTMs, transformers achieve this robustness even with smaller training sets (Section~\ref{sec:other_models}). 
	\item Even though chess is Markovian, the model relies on having access to the whole history, and the performance drops when limiting this access (Section~\ref{sec:limited_history}). 
\end{enumerate*}

\newpage
To summarize, our contributions are to: 

\begin{itemizesquish}
	\itemsep0em 
	\item Propose chess as a testbed for evaluating world state tracking capabilities of language models which can be used for development and analysis of these models.
	\item Show that with the appropriate chess notation, we can probe language models for aspects of the world state using simple prompts (Section~\ref{sec:probing}).
	\item Show that given enough training data, transformer language models can learn to track piece locations and predict legal moves with high accuracy.
	\item Demonstrate that transformer language models are robust to certain changes in input distribution, 
	and that access to world state during training improves performance with small datasets.

\end{itemizesquish}

\section{Chess Preliminaries}
\label{sec:chess}

We represent moves using Universal Chess Interface (UCI) notation, which combines the starting square and the destination square to represent a move.\footnote{For more details see \url{https://en.wikipedia.org/wiki/Universal_Chess_Interface}} 
The move in Figure~\ref{fig:move_notation} is represented as \texttt{f1b5} in UCI where \texttt{f1} indicates the starting square and \texttt{b5} denotes the ending square.
While the SAN notation is the standard choice for gameplay, we prefer UCI %
(see Appendix~\ref{sec:san} for why we pick UCI over SAN). 

For training language models, we first tokenize games represented in UCI notation using a simple regular expression based tokenizer, which considers a board square symbol such as \texttt{b1} as a single token.
This gives us a vocabulary of 77 token types, 
which includes the 64 squares, piece type symbols, and other special symbols (see Table~\ref{tab:model_vocab}).\footnote{In initial experiments we used a delimiter token to indicate move boundary. However, removing it did not degrade performance and made training faster due to reduced sequence length.}
For example, the move sequence ``\pos{e2e4 e7e5 g1f3}" is tokenized to ``\pos{e2}, \pos{e4}, \pos{e7}, \pos{e5}, \pos{g1}, \pos{f3}". We then train an autoregressive language model on these move sequences, using the standard maximum likelihood objective.

\begin{table}[t]
\centering{
\begin{tabular}{llc}
    \toprule
    Type & Examples & Count \\
    \midrule
    Square names & \pos{e4}, \pos{d1} & 64 \\
    Piece type & \pos{P}, \pos{K}, \pos{Q}, \pos{R}, \pos{B}, \pos{N} & \phantom{1}6\\
    Promoted Pawn Piece type & q, r, b, n & \phantom{1}4 \\
    Special symbols & BOS, EOS, PAD & \phantom{1}3 \\
    \midrule
    Total & & 77\\
    \bottomrule
\end{tabular}
}
\caption{Model Vocabulary}
\label{tab:model_vocab}

\end{table}

\begin{table*}
	\centering{
		\begin{tabular}{lll}
			\toprule
			Notation 		& Training 					& Inference \\\midrule
			UCI		 		& \pos{e2}, \pos{e4}, \pos{e7}, \pos{e5}, \pos{g1}, \pos{f3}     						& \pos{e2}, \pos{e4}, \pos{e7}, \pos{e5}, \pos{g1}, \pos{f3} 		\\
			UCI + RAP 15 	& \pos{e2}, \pos{e4}, \pos{P}, \pos{e7}, \pos{e5}, \pos{g1}, \pos{f3} 							& \pos{e2}, \pos{e4}, \pos{e7}, \pos{e5}, \pos{g1}, \pos{f3} 		\\
			UCI + RAP 100 	& \pos{P}, \pos{e2}, \pos{e4}, \pos{P}, \pos{e7}, \pos{e5}, \pos{N}, \pos{g1}, \pos{f3}							& \pos{e2}, \pos{e4}, \pos{e7}, \pos{e5}, \pos{g1}, \pos{f3} 		\\
			UCI + \piecetype 	& \pos{P}, \pos{e2}, \pos{e4}, \pos{P}, \pos{e7}, \pos{e5}, \pos{N}, \pos{g1}, \pos{f3}							& \pos{P}, \pos{e2}, \pos{e4}, \pos{P}, \pos{e7}, \pos{e5}, \pos{N}, \pos{g1}, \pos{f3}		\\
			\bottomrule
		\end{tabular}
	}
		\caption{Token sequences corresponding to the move sequence \pos{e2e4 e7e5 g1f3} for different notations during training and inference. Notice that regardless of the RAP probability used during training, at inference time the token sequences have no piece types.}
	\label{tab:token_seq}

\end{table*}

\section{Language Model Prompts as Board State Probes}
\label{sec:probing}
One attractive property of having a language model trained on chess games represented in UCI notation (as described in the previous section) is that the notation \textit{itself} allows us to probe the trained model's state tracking abilities. In particular, by feeding the trained language model a prefix of a game as a prompt, we can determine --- using the language model's next-token predictions --- what the model understands about the board state implied by this prefix.
For example, consider the prompt ``\pos{\underline{e2e4 e7e5 g1f3 b8c6 d2d4 h7h6} f1},''  where the underlined move sequence leads to the left board state in Figure~\ref{fig:move_notation}. A language model's next-token prediction (after consuming the prompt) can be interpreted as the ending square predicted %
for the bishop at \pos{f1}, which can be used to determine the level of board state awareness of the model. %
If, for instance, the model predicts \pos{g1}, this may indicate that the model does not recognize that the piece type at \pos{f1} is a bishop, as such a move is not possible for a bishop.
If, on the other hand, the model predicts \pos{g2}, that may indicate that the model is not aware that another piece is currently  at \pos{g2}.

\subsection{Randomly Annotated Piece type (RAP)}
\label{sec:rap_board}
While predicting the token representing the ending-square of a move given a prompt allows us to assess the model's state tracking abilities, it also to some extent conflates the model's understanding of the board state with its understanding of chess strategy. If we could easily probe for where the model thinks a piece \textit{currently} is (rather than where it is likely to end up) given a game prefix, this would allow us to more directly probe the model's state tracking abilities. 
In particular, we would like to give a language model a prompt such as ``\pos{e2e4 e7e5 g1f3 b8c6 d2d4 h7h6 \underline{N}}", where \pos{N} represents knight, and expect it to generate a valid starting position for a knight of the correct color. 
While UCI notation does not ordinarily include these piece type tokens, to allow for testing the model with such prompts, 
we propose to randomly include these piece types tokens in moves during training with some fixed probability $p$.
We refer to this strategy as ``randomly annotated piece type'' (RAP) and 
use the nomenclature ``UCI + RAP $p$'' to indicate that with $p\pct$ probability, piece type is part of the move notation during training.
Note that for $p = 0$, the notation reduces to UCI. 

When \emph{testing} with these starting square prediction prompts, we only include piece type for the prompt, not for any moves in the history.
Thus, using RAP during training allows us to probe, at test time, where the model thinks each piece is, given any game history's prefix; by simply providing the desired piece type (e.g., \pos{N}) the model outputs the predicted starting square for a piece of that type.
For example, given the prompt ``\pos{e2e4 e7e5 g1f3 b8c6 d2d4 h7h6 N}", a prediction of \pos{f3} or \pos{b1} shows that the model is aware of where the knights are.%

We also experiment with an ``oracle" variant of RAP where piece types are added both during training and testing. We refer to this notation as ``UCI + \piecetype" where AP stands for ``always piece type".
For our running example the equivalent prompt in this notation would be ``\pos{Pe2e4 Pe7e5 Ng1f3 Nb8c6 Pd2d4 Ph7h6 N}".

In terms of the language modeling training objective, addition of RAP represents a distribution change between training and inference.
Table~\ref{tab:token_seq} illustrates how the use of RAP changes the token sequence during training but not during inference.  
While there's a distribution mismatch, we hypothesize that addition of RAP can aid the model in learning to track the pieces by providing additional supervision which, in turn, can improve language modeling performance as well.

\begin{table*}[t]
\centering
\begin{tabular}{lccc}
\toprule
Task & Prompt Token & Correct Answers (\exactmove) & Correct Answers (\legalmove) \\
\midrule
 End-Actual & \pos{f1} & \{\pos{b5}\} & \{\pos{e2, d3, c4, b5 ,a6} \} \\
End-Other & \pos{f3} & N/A & \{\pos{d2, g1, h4, g5, e5}\} \\
  \midrule
Start-Actual & \pos{B} & \{\pos{f1}\} & \{\pos{f1, c1}\} \\
Start-Other & \pos{N} & N/A & \{\pos{f3, b1}\} \\
\bottomrule
\end{tabular}
\caption{Examples of each probing task, as well as the corresponding exact move (\exactmove) and legal move (\legalmove) correct answers, are shown below. All examples assume the language model was fed the prefix \pos{e2e4 e7e5 g1f3 b8c6 d2d4 h7h6} (see Figure~\ref{fig:move_notation}), and that the actual next move was \pos{f1b5}. While there is only one valid prompt token for both End-Actual and Start-Actual tasks, there are many valid prompt tokens for the other tasks, and we show just one possibility for each. Start-tasks (bottom sub-table) assume the model was trained on games described in UCI+RAP notation.}
\label{tab:tasks}

\end{table*}

\subsection{Board State Probing Tasks}
\label{sec:cloze}
In this subsection we describe the probing tasks introduced above more concretely. %
In each probing task we feed the model a prefix of a game followed by a single prompt token, and the model is evaluated based on the highest probability next-token under the model given this context. We show an example of each probing task in Table~\ref{tab:tasks} (which we further describe below), assuming the model has been fed the move sequence prefix \pos{e2e4 e7e5 g1f3 b8c6 d2d4 h7h6}, %
which is visualized as the left board in Figure~\ref{fig:move_notation}. The actual next move played in the game is \pos{f1b5}, which takes the white bishop at square \pos{f1} to square \pos{b5}, as shown in the right board of Figure~\ref{fig:move_notation}.

\subsection{Ending Square Tasks}
In this set of tasks, the model is given a game prefix and prompted with the starting square of the next move (\pos{f1} in the example of Table~\ref{tab:tasks}). The model's next-token prediction represents its prediction for the ending square of this move,
which
tests the model's ability to track the board state and follow
the rules of chess,
as well as strategic awareness.\footnote{Strategic capabilities of a chess language model are strongly tied to the quality of training games.}  We consider two task variants: %
\begin{enumerate}
	\item \textbf{End-Actual}: Given a move sequence prefix, the model is prompted with the starting square of the actual piece moved next in the game. %
	\item \textbf{End-Other}: Given a move sequence prefix, the model is prompted with the starting square of any piece on the board that can be legally moved according to the rules of chess. 
\end{enumerate}
We evaluate End-Actual predictions in terms of both exact move (\exactmove) accuracy (whether the model predicted the true ending square, \pos{b5} in our running example) and legal move (\legalmove) accuracy (whether the model predicted a legal ending square for the piece starting at the square in the prompt). 
For \legalmove evaluation, we also calculate the R-Precision which is the Precision@R where R is the total number of legal ending squares~\cite{ir-book}. In our running example, there are 5 legal ending squares, and R-Precision will be calculated for the model's top-5 predictions.
\exactmove accuracy evaluation is similar to the typical evaluation of language models on natural language data, while \legalmove is less stringent and focuses on testing just the model's understanding of chess rules and the board state. Note that for End-Other, only \legalmove evaluation is available. See Table~\ref{tab:tasks} for examples.

\subsection{Starting Square Tasks}
In this category of task, the model is again given a game prefix, but prompted with just the piece type of the next move, such as \pos{B} for bishop in the example in Table~\ref{tab:tasks}. The model's next-token prediction thus represents its prediction for where the prompted piece type currently is on the board. This task tests the model's ability to track pieces.\footnote{In certain cases, this task also tests understanding of chess rules. For example, in Figure~\ref{fig:move_notation} only the rook at \pos{h1} can be moved.}
Note that only models which have seen piece types during training, i.e.\ ``UCI + RAP'' models, can actually be tested on this task.
Also, no piece types are used in the game prefix. %
We again have two variants of this task:
\begin{enumerate}
	\item \textbf{Start-Actual}: Given a move sequence prefix, the model is prompted with the piece type of the actual piece moved next in the game. 
	\item \textbf{Start-Other}: Given a move sequence prefix, the model is prompted with the piece type of any piece on the board that can be legally moved according to the rules of chess. %
\end{enumerate}
We again evaluate Start-Actual %
both in terms of \exactmove accuracy (whether the model predicts the starting square of the piece actually moved next in the game), as well as in terms of \legalmove accuracy (whether the model predicts the starting square of a legally movable piece of the given piece type) and \legalmove R-Precision (precision of the model's top-R predictions with respect to all of the R starting squares of legally movable pieces of the given piece type). For Start-Other, only \legalmove evaluation is applicable; see Table~\ref{tab:tasks} for examples.

\section{Experimental Setup}
\label{sec:setup}

\paragraph{Data}
We use the Millionbase dataset which is freely available and has close to 2.9 million quality chess games.\footnote{Download link available at \url{https://rebel13.nl/rebel13/rebel\%2013.html}}
After filtering out duplicate games, games with
fewer than 10 moves, and games with
more than 150 moves (for the complete game to fit into one transformer window), we are left with around 2.5 million games.
From this filtered set we randomly select 200K games for training, 15K games each for dev and test, and another 50K games to create board state probing evaluation sets described in Section~\ref{sec:cloze}.
The dev and test sets are used for perplexity evaluations. 
The dev set perplexity is used for choosing hyperparameters.
From the 200K training set, we create subsets of size 15K and 50K which we refer to as ``Train-S'' and ``Train-M'', while the full training set is referred to as ``Train-L''.
For detailed statistics, see Table~\ref{tab:data_stats} in Appendix.
All the data processing steps requiring chess knowledge, including parsing chess databases, are carried out using python-chess~\citep{python-chess}.

To create the board state probing evaluation sets, we use the 50K games reserved for this task. %
We only consider prompts for non-pawn pieces since the dynamics of pawns are fairly limited.
We ensure that the game prefixes selected are never seen in the training data.
The final evaluation set consists of 1000 instances with prefix length (in number of moves) in the range $51 \le l \le 100$.

\begin{figure*}
	\begin{minipage}{\textwidth}
		\begin{minipage}[b]{0.48\textwidth}
			\centering
			\begin{tabular}{llcc}
				\toprule
				Training Set & Model   & Dev set & Test set \\
				\midrule
				\multirow{2}{*}{Train-S} 
				& UCI 				& 23.6 & 23.6\\
				& UCI + RAP 		& 15.9 & 15.9\\
				& UCI + \piecetype 	& 16.1 & 16.2 \\
				\midrule
				\multirow{2}{*}{Train-M} 
				& UCI 				& 11.6 & 11.6\\
				& UCI + RAP 		& 10.4 & 10.4\\
				& UCI + \piecetype 	& 10.1 & 10.0 \\
				\midrule
				\multirow{2}{*}{Train-L} 
				& UCI 				& \phantom{1}7.7 & \phantom{1}7.7\\
				& UCI + RAP 		& \phantom{1}7.4 & \phantom{1}7.4\\
				& UCI + \piecetype 	& \phantom{1}7.2 & \phantom{1}7.2 \\
				\bottomrule
				
			\end{tabular}
			\captionof{table}{Canonical validation and test set perplexity. By canonical we mean that one move, say \pos{f1b5}, counts as one token.}
			\label{tab:perplexity}
		\end{minipage}
		\hfill
		\begin{minipage}[b]{0.48\textwidth}
			\centering
			\includegraphics[width=\textwidth]{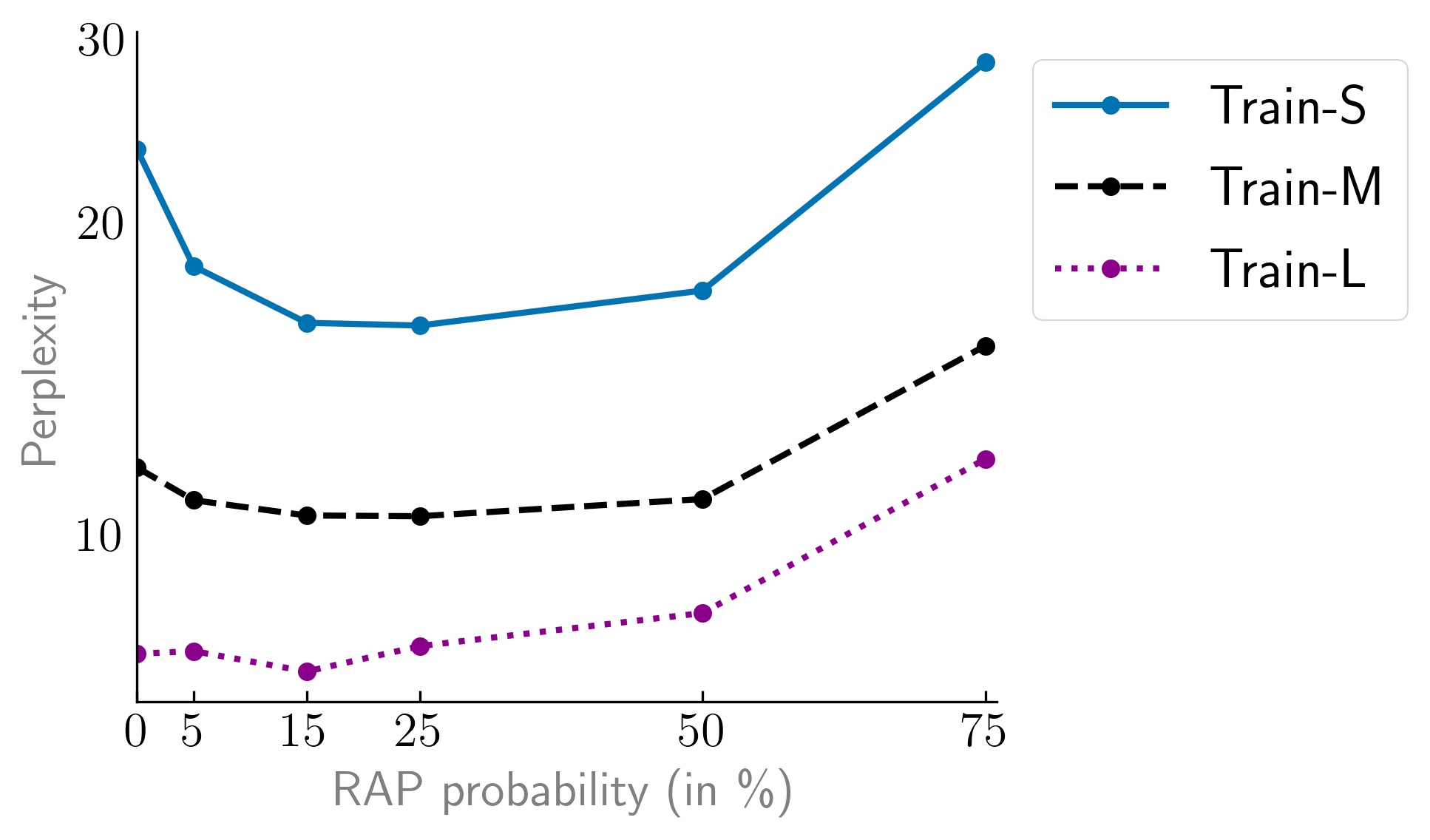}
			\captionof{figure}{Validation set perplexities as a function of RAP probabilities for the different training set sizes. RAP $0$ %
				is %
				the standard UCI notation. 
				RAP $100$ is not shown as perplexities are too high. }
			\label{fig:rap_vals}
		\end{minipage}
	\end{minipage}
\end{figure*}

\paragraph{Model Details}
We use the GPT2-small architecture for our base language model \citep{vaswani2017attention,radford2019language}.
GPT2-small is a 12-layer transformer model with 12 attention heads and an embedding size of 768 dimensions.
The context size of the model is limited to 512, which is sufficient to cover the longest game in our training set.
Note that we only borrow the model architecture; the models themselves are \emph{trained from scratch}.
\footnote{Colab notebook to play chess against the base language model \url{https://github.com/shtoshni/learning-chess-blindfolded/blob/master/GPT2_Chess_Model.ipynb}}

For the UCI + RAP $p$ models, we tune over $p \in \{5, 15, 25, 50, 75, 100\}$ based on %
perplexity on the validation set.
Note that for perplexity evaluation, logits corresponding to piece type tokens are masked out since piece type tokens are only available during training.
We find that $p=25$ performs the best for Train-S and Train-M, while $p=15$ is best for Train-L (Figure~\ref{fig:rap_vals}). %
Larger values of $p$ lead to greater mismatch between training and inference, while smaller values likely do not provide enough training signal.

We also experiment with other transformer and non-transformer models in Section~\ref{sec:other_models}.
Among the transformer models, we experiment with two ``approximate" attention models (i.e., models which approximate the full attention of vanilla transformer models), namely, Reformer \cite{kitaev2020reformer} and Performer \cite{choromanski2021rethinking}.  
We set the number of layers and attention heads to 12 for both 
architectures, as in GPT2-small.
We also train LSTM language models with and without RAP. 
For details on hyperparameters and tuning, see Appendix~\ref{sec:hyperparams}.

\paragraph{Training Details}
Models are trained for 10 epochs with a batch size of 60. Validation is performed %
 at the end of every epoch and training stops whenever the validation loss starts increasing.
For optimization we use Adam \citep{kingma2014adam} with learning rate of $5\times10^{-4}$ and L2 weight decay of $0.01$.
The learning rate is warmed up linearly over the first 10\% of training followed by a linear decay.
To accelerate training, we use mixed precision training~\citep{micikevicius2018mixed}. %
All experiments are carried out using the PyTorch Lightning framework %
built on top of PyTorch \citep{falcon2019pytorch, pytorch}.
We use the transformers library \citep{Wolf2019HuggingFacesTS} for all models\footnote{Reformer implementation in \pos{transformers} library is still a work in progress. The presented results are with the 4.2.2 version.} %
except for the Performer model %
for which we use a popular unofficial implementation.
\footnote{\url{https://github.com/lucidrains/performer-pytorch}}

\section{Results}
We first present language modeling results, where we show significant 
improvements with the addition of RAP (Section~\ref{sec:perplexity_res}). 
Next, we show results on the board state probing tasks for the base language model, where we demonstrate that the 
model trained on the large training set can learn to track pieces and predict legal moves with high accuracy (Section~\ref{sec:state_tracking_res}).
Finally, we present results on the probing task 
with approximate attention transformer architectures and LSTMs,  where we show a performance drop in comparison to the base model with full attention (Section~\ref{sec:other_models}).

\subsection{Language Modeling}
\label{sec:perplexity_res}
Table~\ref{tab:perplexity} presents the perplexity results on the validation and test sets.  
Figure \ref{fig:rap_vals} plots the validation set perplexities as a function of RAP probability for different training set sizes. 
The addition of RAP and \piecetype leads to a decrease in perplexity for all training sizes, particularly for small training sets.
For small training sets, RAP probabilities as high as 50\% can improve the validation perplexity, but for larger training sets, lower RAP probabilities are preferred. 
The reductions in perplexity for RAP are surprising given that the extra tokens added via RAP are not present in the validation and test sets, and thus there is a data distribution shift. %
Models trained with UCI + \piecetype achieve the lowest perplexities on larger training sets. 
Both RAP and \piecetype aid the model in piece tracking, as we will see in later results, and in the case of chess this can significantly improve the language modeling results as well.
Note that for calculating the perplexity of UCI + RAP models, we mask out the logits corresponding to piece type tokens since they are never present during inference.

\subsection{Board State Tracking}
\label{sec:state_tracking_res}
Tables~\ref{tab:results-starting} and~\ref{tab:results-ending} show results when predicting starting squares and ending squares, respectively. 
There are several observations to note. First,  \textbf{transformers can learn to identify where pieces are located}.
This is shown by the \legalmove accuracies in Table~\ref{tab:results-starting}.
UCI + RAP can predict legal starting positions with perfect accuracy and R-Precision. 
However, this capability requires Train-L, and the accuracy drops to 91.3\% for Train-S. 
The gap between UCI + RAP and its ``oracle" counterpart, UCI + \piecetype, also reduces with an increase in training set size with UCI + RAP achieving parity for Train-L.
When asked to identify the location of a piece other than the one selected to be moved next, this accuracy drops only slightly to 99.6\%. 
Typically, the piece location tracking is slightly better for the piece type that is actually moved 
than for other piece types.

The difference between the location of the piece in the exact move (\exactmove) and the location of either piece of the given type (\legalmove) is substantial, at more than 8\% absolute.  
However, this difference relates to chess strategy rather than board state tracking.

\begin{table}[t]
	\setlength\tabcolsep{4pt}
	\centering{
		\begin{tabular}{llccccc}
			\toprule
			&  Notation 	&  \multicolumn{4}{c}{\legalmove} & \exactmove\\
			
			&  &  \multicolumn{2}{c}{Actual} &   \multicolumn{2}{c}{Other} &  \\
			&  &  Acc. & R-Prec. & Acc. & R-Prec. & Acc.\\
			\midrule
			\multirow{2}{*}{S} 	& UCI + RAP& \phantom{1}91.3 & \phantom{1}90.2 & \phantom{1}89.3 & \phantom{1}89.2  & 78.8 \\
			& UCI + AP & \phantom{1}99.2 & \phantom{1}99.1 & \phantom{1}98.8 & \phantom{1}98.8  & 86.9 \\
			\midrule
			\multirow{2}{*}{M} 	& UCI + RAP & \phantom{1}98.2 & \phantom{1}98.0 & \phantom{1}98.6 & \phantom{1}98.7  & 88.0 \\
			& UCI + AP  & \phantom{1}99.9 & \phantom{1}99.8 & 100.0 & 100.0  & 90.2 \\
			\midrule
			\multirow{2}{*}{L} 	& UCI + RAP& 100.0 & 100.0 & \phantom{1}99.6 & \phantom{1}99.5  & 91.8 \\
			& UCI + AP & \phantom{1}99.9 & \phantom{1}99.9 & \phantom{1}99.7 & \phantom{1}99.7  & 91.1 \\
			\midrule
			\multicolumn{2}{l}{Random Legal}  & - & - & - & - & 86.0\\
			\bottomrule
		\end{tabular}
		
	\caption{Accuracies and R-Precisions (\%) for predicting starting squares (``Start-Actual'' and ``Start-Other'' tasks). S, M, L in the first column refer to the training set sizes. 
	}
	\label{tab:results-starting}
	
}
\end{table}
\begin{table}
	\setlength\tabcolsep{4pt}
	\centering{
		\begin{tabular}{llccccc}
			\toprule
			&  Notation 	&  \multicolumn{4}{c}{\legalmove} & \exactmove\\
			
			&  &  \multicolumn{2}{c}{Actual} &   \multicolumn{2}{c}{Other} &  \\
			&  &  Acc. & R-Prec. & Acc. & R-Prec. & Acc.\\
			\midrule
			\multirow{3}{*}{S}
			& UCI  				&  74.0 & 61.1 & 65.5 & 57.7  & 26.7 \\
			& UCI + RAP  		&  88.4 & 75.5 & 80.4 & 72.1  & 33.3  \\
			& UCI + \piecetype 	&  87.0 & 77.0 & 78.8 & 72.3  & 36.1   \\
			\midrule
			\multirow{3}{*}{M}
			& UCI  				&  92.9 & 80.6 & 85.8 & 78.5  & 42.2  \\
			& UCI + RAP  		&  94.9 & 82.2 & 87.9 & 78.0  & 45.9  \\
			& UCI + \piecetype 	&  94.7 & 82.4 & 88.3 & 79.1  & 47.3   \\ 
			\midrule
			\multirow{3}{*}{L}
			& UCI  				&  97.7 & 85.6 & 91.9 & 83.8  & 52.0   \\
			& UCI + RAP  		&  97.0 & 86.1 & 93.1 & 83.9  & 54.7  \\
			& UCI + \piecetype 	&  98.2 & 87.3 & 95.2 & 86.3  & 56.7  \\
			\midrule
			\multicolumn{2}{l}{Random Legal} & - & - & - & - & 19.6 \\
			\bottomrule
			
		\end{tabular}
	\caption{Accuracies and R-Precisions (\%) for predicting ending squares (``End-Actual'' and ``End-Other'' tasks). S, M, L in the first column refer to the training set sizes.
	}
	\label{tab:results-ending}
	
	}
\end{table}

Second, \textbf{transformers can learn to predict legal moves}.
This is shown by the \legalmove accuracies in  Table~\ref{tab:results-ending}, for which both UCI and UCI + RAP exceed 97\% accuracy. 
However, while the top predictions of the models have high accuracy, their ability to predict all legal moves is significantly lower, with R-precision of about 85\%. 
This is to be expected, since the model is trained on only actual games, where the emphasis is on ``meaningful" moves rather than any legal move. 
Due to similar reasons, there's a significant drop in performance when predicting ending squares for starting squares other than the one in the actual game. 
	The ``other" starting square would, by design, have legal continuations, but lack any ``meaningful" ones 	(see examples in Appendix \ref{sec:app_error_analysis}).

We find consistent gains in almost all metrics with the addition of RAP during training, with the gains being particularly impressive for small training sets. Thus, not only are the transformers robust to distribution shift due to RAP (available only during training), they are in fact able to utilize this additional information. Error analysis of illegal predictions shows that the addition of RAP improves piece tracking related errors (Appendix~\ref{sec:error_analysis}).  

The relatively low ExM accuracies of the models can be attributed to the inherent difficulty of the task.   
Randomly selecting an ending square from all legal ending squares 
has an accuracy of only around 20\%, implying that on average there are roughly 5 legal choices, which might explain the difficulty of the task.  

\begin{table}[t]
\centering{
		\setlength\tabcolsep{3.1pt}
		\begin{tabular}{llccccc}
			\toprule
			&  Model 	&  \multicolumn{4}{c}{\legalmove} & \exactmove\\
			
			&  &  \multicolumn{2}{c}{Actual} &   \multicolumn{2}{c}{Other} &  \\
			&  &  Acc. & R-Prec. & Acc. & R-Prec. & Acc.\\
			\midrule
			\multirow{6}{*}{S}
			& GPT2  				&  74.0 & 61.1 & 65.5 & 57.7  & 26.7  \\
			& GPT2 ($w=50$)    			& 69.5 & 57.4 & 60.4 & 53.2  & 23.1  \\
			
			& Reformer				&  71.0 & 57.2 & 61.5 & 53.5  & 24.8 \\
			& Performer				&  65.4 & 54.3 & 57.9 & 49.5  & 20.5  \\
			& LSTM 					&  60.2 & 51.0 & 52.5 & 46.4  & 20.9  \\
			& LSTM + RAP 			&  59.5 & 50.5 & 52.4 & 46.0  & 21.9 \\
			
			\midrule
			\multirow{6}{*}{M}
			& GPT2  				&  92.9 & 80.6 & 85.8 & 78.5  & 42.2   	\\
			& GPT2 ($w=50$)  			&  86.0 & 74.9 & 80.9 & 71.3  & 35.8  	\\
			& Reformer 				&  86.4 & 73.2 & 76.6 & 68.6  & 32.4 	\\
			& Performer 			&  89.2 & 76.3 & 80.5 & 71.5  & 36.0   	\\
			& LSTM  				&  73.8 & 61.6 & 67.2 & 59.8  & 32.0   	\\
			& LSTM + RAP 			&  77.5 & 64.9 & 69.7 & 61.7  & 32.1  	\\ 
			\midrule
			\multirow{6}{*}{L}
			& GPT2  				&  97.7 & 85.6 & 91.9 & 83.8  & 52.0  	\\
			& GPT2 ($w=50$)    		&  95.8 & 84.5 & 90.5 & 82.7  & 51.6  	\\
			& Reformer 				&  88.0 & 74.9 & 77.0 & 68.1  & 33.5 	\\
			& Performer				&  95.8 & 84.5 & 90.5 & 82.7  & 51.6  	\\
			
			& LSTM  				&  93.4 & 79.5 & 86.1 & 76.0  & 45.2  	\\
			& LSTM + RAP 			&  92.8 & 80.4 & 87.3 & 77.1  & 46.0	\\
			\bottomrule
			
		\end{tabular}
		
	\caption{Accuracy and R-Precision (\%) for predicting ending squares (``End-Actual'' and ``End-Other'' tasks)  with varying attention window sizes. 
		LSTM + RAP refers to LSTM  trained with UCI + RAP.
	}
	\label{tab:results-ending-window}
	
	}
\end{table}

\subsection{Compressing the Game History}
\label{sec:other_models}
The base transformer language model, based on GPT2, attends to the entire history (i.e., it uses ``full attention"), which results in complexity quadratic in the length of the sequence. We might wonder whether attending to this entire history is necessary for the impressive state tracking performance observed in the %
previous section.
We accordingly 
explore models that do not attend to the entire history in Table \ref{tab:results-ending-window}. %

We first experiment with a variant of the GPT2 model that limits its attention to a window of only the 50 most recent tokens (``GPT2 $(w=50)$''). In Table \ref{tab:results-ending-window} we see worse performance for this model across data sizes, but especially for small- and medium-sized datasets. 

In Table~\ref{tab:results-ending-window} we also consider a language model based on the LSTM~\citep{hochreiter1997long}, which considers only its current hidden state and cell state in making its predictions, and does not explicitly attend to the history. %
Here we find an even more significant drop in performance, in all settings. (Interestingly, we also find that training LSTM language models on sequences with RAP improves performance, but only for larger training sets; transformer language models generally improve when trained with RAP data). 

The results of GPT2 $(w = 50)$ and of the LSTM language model suggest that attending to the full game history is, unsurprisingly, useful for board state tracking in chess. This finding further suggests that the task of board state tracking in chess can serve as an excellent testbed for recently proposed transformer variants~\citep[\textit{inter alia}]{kitaev2020reformer,katharopoulos20,choromanski2021rethinking} that attempt to make use of long histories or contexts, but \textit{without} incurring a quadratic runtime.

\subsubsection{Approximate Attention Transformers}
\label{sec:limited_history}

We experiment with the recently proposed Reformer~\cite{kitaev2020reformer} and Performer~\cite{choromanski2021rethinking} architectures. Reformer replaces the ``full attention" with attention based on locality-sensitive hashing, while Performer approximates the ``full attention" with random features.\footnote{In practice, these models often use a combination of the proposed approximate global attention and simple local attention (for details see Appendix~\ref{sec:hyperparams}).}

The results, in Table~\ref{tab:results-ending-window}, suggest that the Performer generally outperforms the Reformer, except in the small dataset-setting. Furthermore, we find that neither of these architectures significantly outperforms the GPT2 $(w = 50)$ baseline, except for Performer in the medium-sized data setting. 
These models do, however, typically outperform the LSTM models. 
These results demonstrate the challenge of modeling chess with an approximate attention. 
We hope that future work will use this task as a way of benchmarking more efficient transformer architectures. %

\section{Related Work}
\paragraph{Simulated Worlds.} %
There have been several prior efforts in relating simulated worlds to natural language. 
The bAbI framework simulates a world modeled via templates to generate question answering tasks \citep{weston2015aicomplete}. 
The recent TextWorld framework facilitates generating, training, and evaluating interactive text-based games \citep{cote18textworld}. 
\citet{hermann17grounded} and \citet{hill17understanding} develop and use 3D world simulations for learning grounded language.
These efforts are similar to our work in the sense that the true world state is, by construction, available, but our setup differs in that it provides a natural way of probing the state tracking of a model trained with an LM objective.

\paragraph{Cloze Tasks for Natural Language Models.}
There has been a great deal of work on cloze tasks for evaluating natural language models~\citep{hermann2015cnn, hill2016cbt}. 
These tasks range from testing general text understanding~\citep{paperno-etal-2016-lambada} to targeting particular aspects of natural language, such as commonsense/pragmatics \citep{mostafazadeh-etal-2016-corpus, ettinger2020bert}, narrative understanding \citep{mostafazadeh-etal-2017-lsdsem}, and factual knowledge \citep{petroni-etal-2019-language}.
Creating these tasks often requires human curation, and the evaluation is typically limited to exact match.\footnote{Automated cloze tasks without human filtering can yield instances which even humans can't answer \citep{hill2016cbt}.}  
Our proposed tasks are a form of cloze tasks, but can be precisely 
automated so that they require no human curation, and can be evaluated at a fine-grained level.

\paragraph{Probing.}
One of the goals of this work is to probe the language model's board state tracking capability.
A typical solution used by prior work is to train a probing model on top of a pretrained model  
\citep{ettinger-etal-2016-probing,Alain2017UnderstandingIL, adi17probing, tenney2019probing,hewitt-liang-2019-designing}. %
This setup is time-consuming as it requires training probing models for all tasks. 
Moreover, the complexity of the probing model can also affect the conclusions \citep{pimentel-etal-2020-information}. 
In our case, by using an appropriate choice of notation, probing for board state can be accomplished via simple prompts (Section ~\ref{sec:probing}). 

\paragraph{Deep Learning for Chess.}
Deep networks have been used in prior work to predict the next move given the true game state~\cite{david16deepchess, Oshri2015PredictingMI}.
For example, using only self-play and the rules of chess, AlphaZero achieves superhuman performance starting from random play~\citep{silver18general}.
The focus of this prior work is the quality of game play given the true board state, while we use chess as a testbed for evaluating a language model's board state tracking capability.
Recently there has also been work focusing on transformer language models for chess \citep{presser2020chess,cheng2020chess,noever2020chess}. 
This work is similar to ours in the sense that the input is limited to the move sequence without the true board state, but the focus is again the quality of game play rather than the model's awareness of the underlying state.

\section{Conclusion}
We propose the game of chess as a testbed for evaluating how well language models capture the underlying world state. 
We show that with an appropriate choice of chess notation, a language model can be probed for different aspects of the board state via simple prompts.
The simple and precise dynamics of chess allow for (a) training models with varying amount of explicit state, and (b) evaluating 
model predictions at a fine-grained level.
Results show that transformer language models are able to track the board state when given enough
data, but with limited data, %
providing access to board state information during training can yield consistent improvement. 

 \paragraph{Wider Implications for Natural Language Processing.}
Our results shed light on the following properties of transformers: (a) they are robust to RAP-like changes in input distribution, and (b) for high performance the models require access to the entire context, as well as large training sets (Section~\ref{sec:limited_history}). 
Future work can use the first finding to introduce the world state, or more specifically the output of linguistic analyzers such as coreference,  via RAP-like tokens during pre-training and fine-tuning of transformers. 
RAP-like tokens can also be used for debugging/diagnosing a model's understanding, similarly to the starting square prediction tasks. 
The second finding implies that the proposed benchmark can guide the search for new transformer architectures that are adept at understanding long text, and that can learn from small training sets.  
The proposed framework allows for probing and understanding new architectures that address these challenges.

\section*{Acknowledgements}

We thank Ed Schr\"{o}der for permitting us to use the Millionbase database for this project.
We thank Allyson Ettinger and colleagues at TTI Chicago for their valuable feedback. 
This material is based upon work supported by the National Science Foundation under
Award No. 1941178. 

\bibliography{0-main}

\clearpage
\newpage
\appendix
\newpage

\section{SAN Notation}
\label{sec:san}
\paragraph{Standard Algebraic Notation (SAN)} combines the piece type moved and the destination square to denote a move.\footnote{For more details see \url{https://en.wikipedia.org/wiki/Algebraic_notation_(chess)}}
For example, the move in Figure~\ref{fig:move_notation} is represented as \texttt{Bb5} in SAN where \texttt{B} represents the piece type bishop and \texttt{b5} represents the destination square.

\paragraph{Standard Algebraic Notation (SAN) Ambiguity}
SAN notation doesn't use the starting square of the piece in its move representation.
This limits the ability to prompt a SAN-based language model with specific piece type instances.
For example, given the prompt ``\texttt{\underline{e4 e5 Nf3 Nc6 d4 h6} B}'' (the underlined move sequence leads to the left board state in Figure~\ref{fig:move_notation}), it's not clear whether the token \texttt{B} refers to the bishop at \texttt{f1} or \texttt{c1}. Due to this limitation on the specificity of probing queries, we do not use SAN for our experiments.

\section{Model Vocabulary}

Table~\ref{tab:model_vocab} shows the vocabulary used. 
No delimiter token is used to denote the move boundary. 
Tokens of promoted pawn piece type are used when a pawn gets promoted.
For example, \pos{e7e8q} denotes the move where a pawn from e7 moves to e8 and becomes a queen.

\section{Effect of Model Size}

In this section, we present results for training larger transformer models to evaluate the impact of increase in model size with increase in training set size.

\begin{table}[t]
\caption{Accuracy and R-Precision (\%) for predicting ending squares (``End-Actual'' and ``End-Other'' tasks)  for different model sizes.
	S, M, L in the first column refer to the training set sizes.
	 GPT2-small = \{12 layers, 12 heads, 768 embedding size\}; GPT2-intermediate = \{16 layers, 12 heads, 768 embedding size\}; and GPT2-medium = \{24 layers, 16 heads, 1024 embedding size\}.
}
\label{tab:results-ending-size}
    \centering{
\setlength\tabcolsep{4pt}
\begin{tabular}{llccccc}
	\toprule
    	&  Model 	&  \multicolumn{4}{c}{\legalmove} & \exactmove\\
    
    &  &  \multicolumn{2}{c}{Actual} &   \multicolumn{2}{c}{Other} &  \\
    &  &  Acc. & R-Prec. & Acc. & R-Prec. & Acc.\\
    \midrule
\multirow{3}{*}{S}
& GPT2-small  		&  74.0 & 61.1 & 65.5 & 57.7  & 26.7  \\
& GPT2-inter.  	&  72.3 & 60.7 & 64.5 & 58.6  & 24.8  \\
& GPT2-med.  		&  67.8 & 58.2 & 62.5 & 55.7  & 24.5  \\
\midrule
\multirow{3}{*}{M}
& GPT2-small  		&  92.9 & 80.6 & 85.8 & 78.5  & 42.2  \\
& GPT2-inter.  	&  92.9 & 81.8 & 84.8 & 77.8  & 41.5  \\
& GPT2-med.  		&  93.7 & 81.8 & 86.2 & 77.1  & 41.7  \\
\midrule
\multirow{3}{*}{L}
& GPT2-small  		&  97.7 & 85.6 & 91.9 & 83.8  & 52.0  \\
& GPT2-inter.  	&  97.5 & 86.6 & 94.7 & 85.2  & 54.0  \\
& GPT2-med.  		&  98.2 & 87.4 & 94.6 & 85.8  & 57.0  \\
\bottomrule

\end{tabular}
}
\end{table}

Table~\ref{tab:results-ending-size} presents results with transformer models of sizes varying from GPT2-small to GPT2-medium.
We also introduce a new configuration, referred to as GPT2-intermediate, which serves as an intermediate between GPT2-small and GPT2-medium.
For Train-S, GPT2-small outperforms both GPT2-intermediate and GPT2-medium on almost all evaluations.
However, with increasing in training data,  GPT2-intermediate and GPT2-medium are are able to
outperform GPT2-small on most evaluations. 

These results are along the expected lines of larger training sets alleviating the overfitting problem with larger models~\citep{kaplan2020scaling}.  
Note that we stick with the default GPT2 configuration for all our experiments. Tuning the regularization hyperparameters such as dropout, can further improve results for bigger models trained with small training sets.

\begin{figure*}
\begin{minipage}{\textwidth}
	\begin{minipage}[b]{0.5\textwidth}
		\setlength\tabcolsep{4pt}
		\begin{tabular}{lcc}
			\toprule
			Split & \# of games (in $10^3$) & \# of moves (in $10^6$)\\ %
			\midrule
			Train-S & 	\phantom{1}15 	& \phantom{1}1.1\\ %
			Train-M & 	\phantom{1}50 	& \phantom{1}3.7\\ %
			Train-L & 	200 			& 15.0\\ 			 %
			Dev 	& 	\phantom{1}15 	& \phantom{1}1.1\\ %
			Test 	& 	\phantom{1}15 	& \phantom{1}1.1\\ %
			\bottomrule
		\end{tabular}
		\captionof{table}{Statistics of the language modeling data.}
		\label{tab:data_stats}
	\end{minipage}
	\hfill
	\begin{minipage}[b]{0.48\textwidth}
		\centering
		\includegraphics[width=0.8\textwidth]{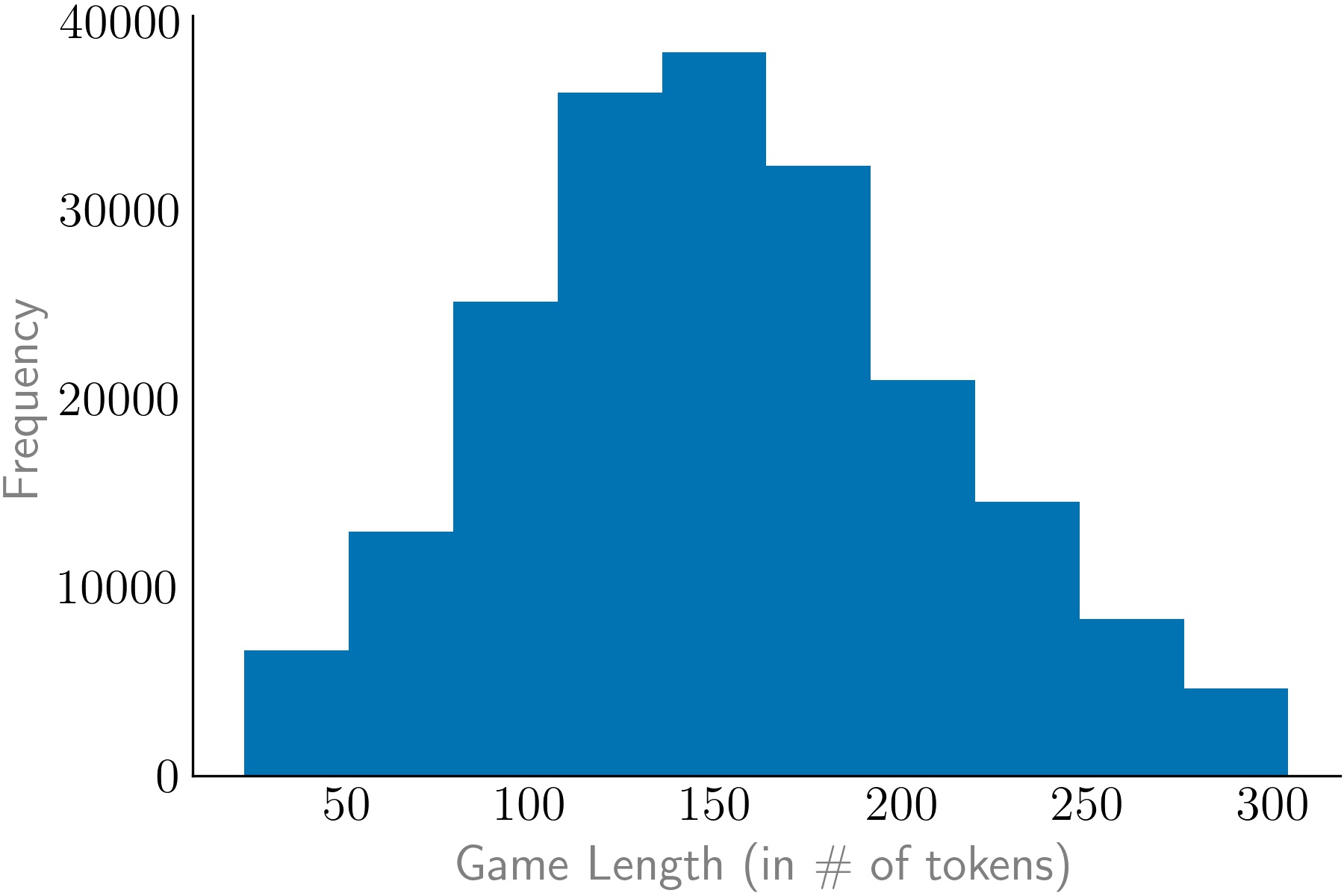}
		\vspace{-0.1in}
		\captionof{figure}{Histogram of tokenized game lengths for Train-L.}
		\label{fig:hist_len}
	\end{minipage}
\end{minipage}
\end{figure*}

\begin{figure*}
	\begin{minipage}{\textwidth}
		\begin{minipage}[b]{0.48\textwidth}
			\centering
			\setlength\tabcolsep{4pt}
			\begin{tabular}{lccc}
				\toprule
				Piece type & End/Start-Actual & End-Other & Start-Other\\ %
				\midrule
				Rook (\pos{R}) & 	358 	& 273     & 197\\ 
				Knight (\pos{N}) & 	144 	& 136 & 126\\  
				Bishop (\pos{B}) & 	164 			& 170 & 161\\
				Queen (\pos{Q}) 	& 	204 	& 103  & 129 \\ 
				
				King (\pos{K}) & 	130 	& 318  & 387\\
				\midrule
				Total & 1000 & 1000 & 1000 \\\bottomrule
			\end{tabular}
			\captionof{table}{Piece type counts for ending square prediction prompts.}
			\label{tab:probe_data_stats}
		\end{minipage}
		\hfill
		\begin{minipage}[b]{0.48\textwidth}
			\centering
			\includegraphics[width=0.8\textwidth]{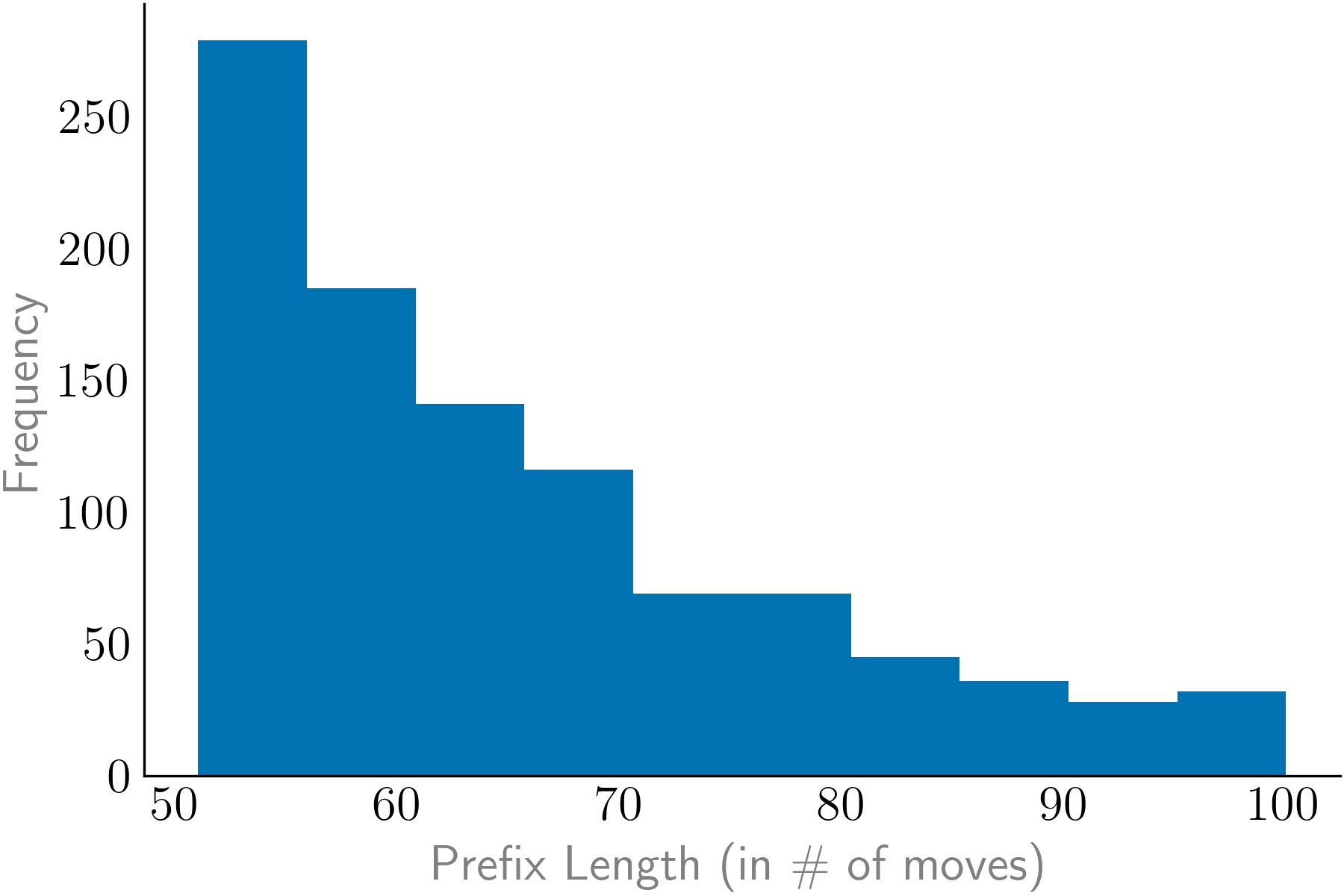}
			\vspace{-0.1in}
			\captionof{figure}{Histogram of prefix lengths of board state prompts.}
			\label{fig:hist_prompt}
		\end{minipage}
		\end{minipage}
\end{figure*}

\section{Data Statistics}
\label{sec:data_stats}
Table~\ref{tab:data_stats} presents the statistics of the language modeling dataset used. The average game length for all splits is around 75 moves.
Figure \ref{fig:hist_len} presents the histogram of lengths of tokenized UCI games in Train-L. 

Table~\ref{tab:probe_data_stats} presents the piece type counts for the different board state prompts. All the prompts have the same game prefix i.e. the previous moves, though, the move prefix is different - starting square of the move is used for the ending square predictions while the piece type used for the move is used for the starting square prediction. As the game prefix is the same, End-Actual and Start-Actual use the same piece type for each prompt. 
For the End-Other task, we pick a random starting square among all starting squares from which a legal move can be made, except the starting square used for the actual move.
For the Start-Other task, we pick a random piece type among all piece types which can be legally moved, except the piece type which is actually moved in the game.
The different strategies for picking the random starting square and random piece type explains the different piece type distributions for End-Other and Start-Other. Figure~\ref{fig:hist_prompt} shows the histogram of length of game prefixes (in number of moves) used in board state prompts.

\section{Model Hyperparameters and Training time}
\label{sec:hyperparams}
Table~\ref{tab:hyperparam} presents the hyperparameters used for the different models. For the base language model based on GPT2-small we use the default hyperparameters. For other baselines we perform separate hyperparameter grid search for Train-S and Train-M, and use the Train-M hyperparameters for Train-L. 
Only exception to this rule is the Reformer model, which we found particularly difficult to train, for which we explain the details next.

Reformer model uses a combination of local and LSH-based self attention layers. We borrow the attention layer configuration used for enwiki8 experiments in the original paper. \footnote{\url{https://cdn.huggingface.co/google/reformer-enwik8/config.json}} 
For both the local and LSH attention, we use a chunk length of 50 tokens - the model divides the sequence into chunks with the causal attention limited to tokens within a chunk and one before.
The transformers library implementation suggests not pre-specifying the number of hash buckets. The implementation sets the number of buckets on the fly based on the sequence length, which in this case it sets to 8 hash buckets. The original paper experiments with the number of hashing rounds and shows consistent improvement with more hashing rounds. However, we didn't find that to be the case, and hyperparameter tuning sometimes preferred lower number of hashing rounds. 
We found it particularly difficult to train the model on Train-L where the training loss started increasing after only a couple of epochs which triggered early stopping. To alleviate this: (a) we experimented with a different learning rate decay mechanism, namely, the inverse square root decay schedule which lead to slightly better final results  \footnote{\url{https://fairseq.readthedocs.io/en/latest/_modules/fairseq/optim/lr_scheduler/inverse_square_root_schedule.html}}, and (b) perform a separate hyperparameter tuning for Train-L.  
Note that all other experiments use the learning rate schedule described in Section~\ref{sec:setup} and use the hyperparameters for Train-M.  

\paragraph{Training Time}
Experiments with transformers take around 4 hrs for Train-S, less than 10 hrs for Train-M, and less than 24 hrs for Train-L on a single GeForce RTX 2080 Ti. For LSTMs it takes less than 2 hrs for Train-S, less than 4 hrs for Train-M, and less than 8 hrs for Train-L on a single GeForce RTX 2080 Ti.

\begin{table*}
	
\centering{
	\caption{Hyperparameters used for the different models.
		Bold values are selected for all the training set sizes, otherwise, training set specific hyperparameter values are specified via parenthesis.
	}
	\label{tab:hyperparam}
	\begin{tabular}{lllll}
		\toprule
	  Hyperparameters&	GPT2	 		& 	LSTM	  	& 	Reformer		&	 Performer \\\midrule
\# of layers  &  	12		& 	3 (S), 4 (M, L), 5				& 12 & 12 \\
\# of attention heads  & 12		& 	0	&  12 & 12\\
Embedding size 	& 768	& \textbf{768}, 1024 	& 768 & 768 \\
Hidden size		& 768	& 768, \textbf{1024}	& 768 & 768 \\
Dropout probability & 0.1	& 0, 0.1, \textbf{0.2}, 0.5 & 0.05 (0 for LSH attn.) & 0.1\\
\# of hash buckets	  & - & - & 8 & -\\
\# rounds of hashing  & - & - & 1 (L), 2 (S), 4 (M) & - \\
Axial position shape  & - & - & [14, 25] &- \\
Axial position embedding size  & - & - & [256, 512] &- \\
Generalized attention & - & - & - & \textbf{Yes}, No \\
Feature redraw frequency  			& - & - & - & 1000 \\
\# of local attention heads 		& - & - & -  & 0 (M, L), 6 (S) \\
Local/LSH attn chunk size 				& - & - & 50 & - 				\\
Local attn window size 				& - & - & - & 50 					\\\midrule
\# of parameters (in millions) 	   	& 85	& 24 (S)/32 (M, L)  & 83 & 86\\
				
\bottomrule

\end{tabular}
}
\end{table*}
\begin{figure*}[t]
\begin{subfigure}{0.3\textwidth}
   \includegraphics[width=\linewidth]{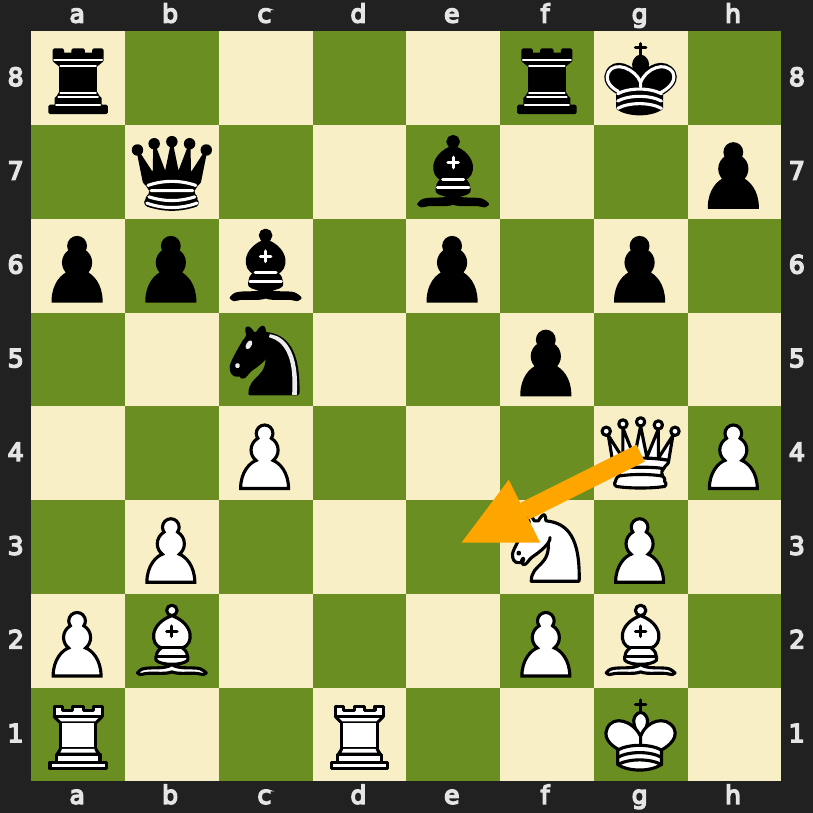}
   \caption{\emph{Syntax}: Queen can move like all other piece types except for knight.} \label{fig:error_syntax}
\end{subfigure}
\hspace*{\fill}
\begin{subfigure}{0.3\textwidth}
   \includegraphics[width=\linewidth]{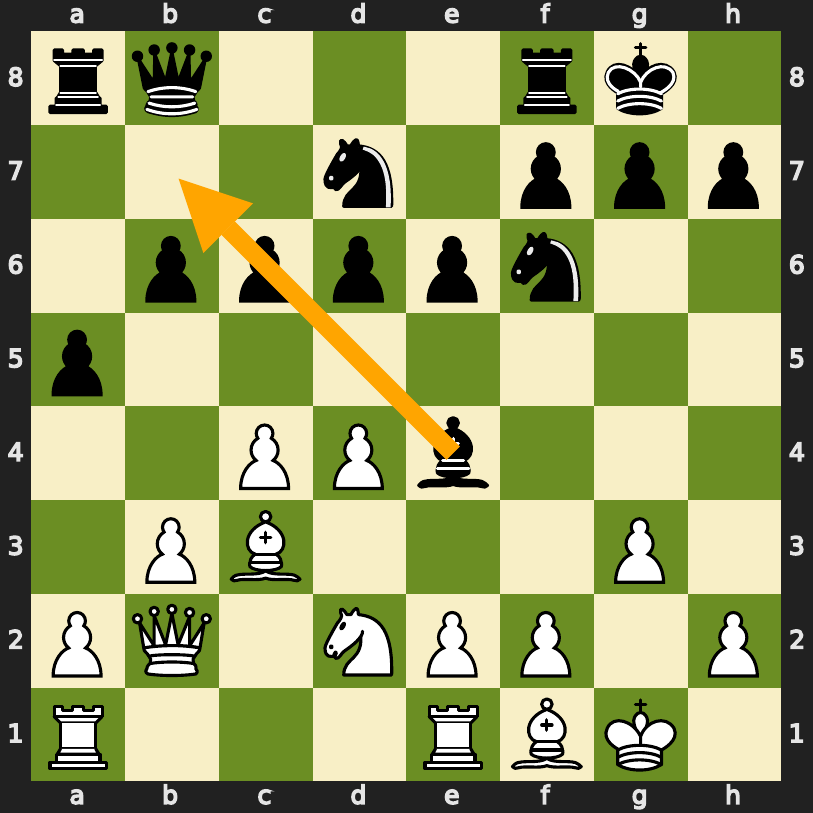}
   \caption{\emph{Path Obstruction}: The pawn at \pos{c6} is blocking the bishop.} \label{fig:error_path}
\end{subfigure}
\hspace*{\fill}
\begin{subfigure}{0.3\textwidth}
   \includegraphics[width=\linewidth]{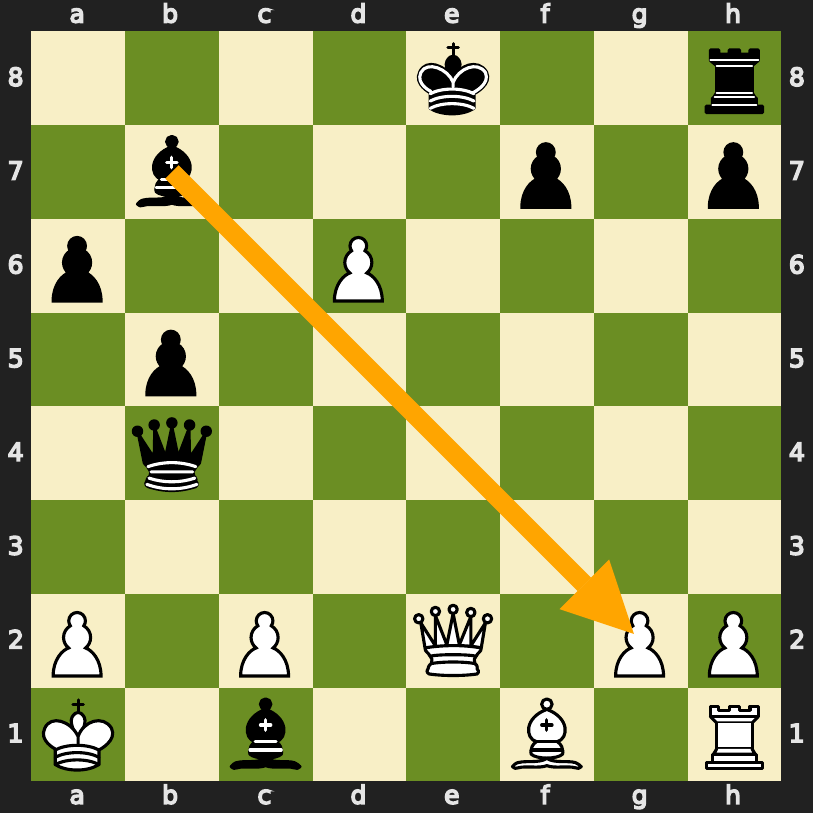}
   \caption{\emph{Pseudo Legal}: The black king remains in check.} \label{fig:error_pseudo}
\end{subfigure}
\caption{Instances of the three prominent categories of illegal ending square predictions.}
\label{fig:error_categories}
\end{figure*}

\begin{table*}[ht]
	\centering{
		\caption{Error counts for ending square prediction.}
		\label{tab:error_analysis}
\begin{tabular}{llcccccc}
	\toprule
	& \multirow{2.5}{*}{Model}   &  \multicolumn{3}{c}{Actual}      &   \multicolumn{3}{c}{Other}    \\
	\cmidrule(lr){3-5}\cmidrule(lr){6-8}
	&        & Syntax & Path Obst. & Pseudo Leg.  & Syntax & Path Obst. & Pseudo Leg.\\
	
	\midrule
	
	\multirow{4}{*}{Train-S} 
	& UCI 				& 168 & 48 & 40 & 173 & \phantom{1}90 & \phantom{1}80 \\ 
	& UCI + RAP			& \phantom{1}20 & 58 & 38 & \phantom{1}17 & \phantom{1}96 & \phantom{1}81\\
	& UCI + \piecetype	& \phantom{11}1 & 99 & 29 & \phantom{11}3 & 126 & \phantom{1}81\\
	& Performer 		& 235 			& 56 & 53 & 243 & \phantom{1}70 & 106\\
	
	\midrule
	
	\multirow{4}{*}{Train-M} 
	& UCI 				& \phantom{1}16 & 30 & 25 & \phantom{1}15 & \phantom{1}54 & \phantom{1}72 \\ 
	& UCI + RAP			& \phantom{11}3 & 30 & 18 & \phantom{11}7 & \phantom{1}56 & \phantom{1}55\\
	& UCI + \piecetype	& \phantom{11}0 & 36 & 17 & \phantom{11}3 & \phantom{1}59 & \phantom{1}53 \\
	& Performer			& \phantom{1}41 & 27 & 40 			& \phantom{1}42 	& \phantom{1}45 & 108\\

	\midrule
	
	\multirow{4}{*}{Train-L} 
	&  UCI 				& \phantom{11}1 & 10 & 12 & \phantom{11}4 & \phantom{1}26 & \phantom{1}49 \\ 
	& UCI + RAP			& \phantom{11}0 & 19 & 11 & \phantom{11}3 & \phantom{1}29 & \phantom{1}36 \\
	& UCI + \piecetype 	& \phantom{11}0 & 13 & \phantom{1}5 & \phantom{11}3 & \phantom{1}13 & \phantom{1}31\\
	& Performer			& \phantom{11}8 & 18 & 16 & \phantom{11}9 & \phantom{1}23 & \phantom{1}63 \\
	\bottomrule
\end{tabular}
	}
\end{table*}

\section{Error Analysis}
\label{sec:error_analysis}

In this section we analyze errors %
on the ending square prediction task. 
Incorrect predictions for this task %
can be (exhaustively) categorized into four categories:

\begin{itemizesquish}
	\itemsep0em 
	\item {\em Unreachable}: The predicted ending square cannot be reached %
	by any possible piece type %
	at the starting square regardless of the board state. 
	\item {\em Syntax}: The predicted ending square cannot be reached %
	by the piece type present at the starting square regardless of the board state. This error indicates failure at tracking the piece type present at the starting square. 
	\item {\em Path Obstruction}: The predicted ending square cannot be reached 
	because there are other pieces obstructing the %
	path. This error indicates failure at tracking other pieces on the board or a lack of %
	understanding that for all piece types except the knight, the path %
	must be clear. %
	For example, in Figure~\ref{fig:error_path}, the pawn at \pos{c6} blocks the bishop's move from \pos{e4} to \pos{b7}.
	\item {\em Pseudo Legal}: 
	The move is illegal because the moving player's king is in check at the end of the move. 
\end{itemizesquish}
Table~\ref{tab:error_analysis} shows error counts for the ending square prediction task. 
For brevity we omit unreachable errors since they are rare ($< 5$ for all models).

Errors across all categories decrease with more training data. For syntax errors this reduction is particularly dramatic, decreasing by roughly an order of magnitude when moving from Train-S to Train-M. %
In contrast, both path obstruction and pseudo legal errors decline more gradually.
Determining whether a path is blocked or if the king is in check requires a computation involving multiple piece locations which all need to be computed from the move  history. 
These trends suggest that identifying the piece type at a starting square 
requires data but is learnable, while keeping track of all \emph{other} pieces  on the board remains challenging even with large training sets.

UCI + RAP %
consistently outperforms %
UCI in syntax errors, %
the differences being largest for the small training sets. This validates our hypothesis that RAP can aid the model in piece tracking (Section~\ref{sec:rap_board}). Across other error categories we don't see consistent trends, suggesting piece tracking improvements do not necessarily translate to other error categories. The Performer generally makes more errors than the transformers, especially in the syntax category. The partial attention in the Performer may be limiting its ability to attend to the most relevant prior positions to determine the piece type at the given starting square. 

Predicting ending squares for the actual move made (``Actual'') is easier than for a randomly chosen legal move (``Other''). However, the syntax errors are comparable between the two settings, while there are many more path obstruction and pseudo legal errors for the Other instances. 
	The higher error rate for these categories could be because:
	\begin{itemizesquish}
		\item Avoiding path obstruction and check are difficult functions to learn and may therefore be being ``mimicked'' from training data rather than being learned as a general algorithmic function.
		\item The model is trained on only actual games with emphasis on meaningful moves rather than legal moves. We observe that some of the Other instances lack any ``meaningful" continuations (Appendix~\ref{sec:app_error_analysis}).
		\item There's a distribution shift between piece types moved in actual moves vs randomly chosen legal moves. 
		For example, the End-Actual task has only about 13\% prompts for moves made by king in comparison to the 33\% for the End-Other task (Appendix~\ref{sec:data_stats}).  We find that moves made by king have a higher chance of resulting in pseudo legal errors in comparison to other piece types (Appendix~\ref{sec:pseudo_legal}).  
	\end{itemizesquish}

\subsection{Detailed Error Analysis}
\label{sec:app_error_analysis}
\begin{figure*}[!ht]
	\centering{
		\hspace*{\fill}
\begin{subfigure}{0.43\textwidth}
   \includegraphics[width=\linewidth]{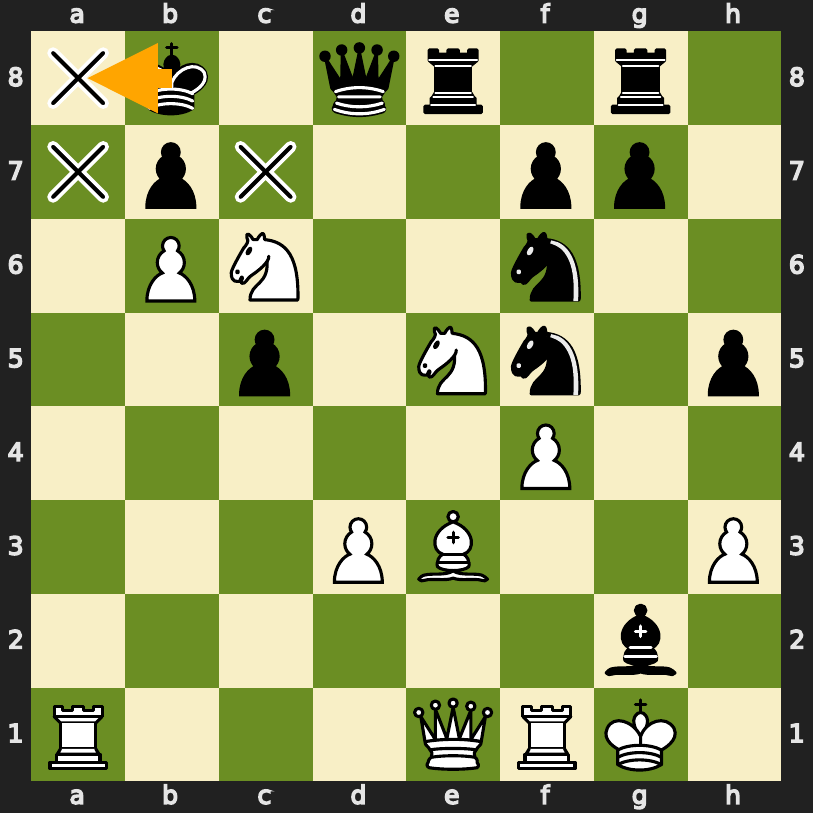}
   \caption{\emph{Check + King}: Black king is in check and the predicted ending square is already covered by the white rook on \pos{a1}.} 
   \label{fig:error_check_king}
\end{subfigure}
\hspace*{\fill}
\begin{subfigure}{0.43\textwidth}
	\includegraphics[width=\linewidth]{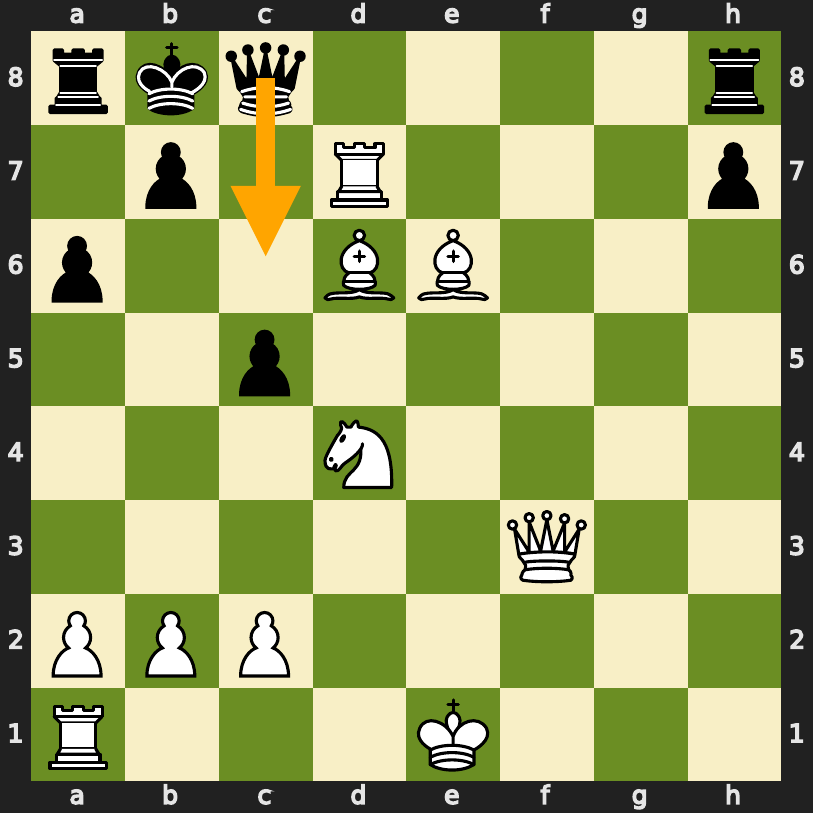}
	\caption{\emph{Check + Other}: Black king is in check and the only legal move for the black queen is \pos{c7} but the model predicts \pos{c6}.} 
	\label{fig:error_check_no_king}
\end{subfigure}
\hspace*{\fill}
\\[1em]
\hspace*{\fill}
\begin{subfigure}{0.43\textwidth}
   \includegraphics[width=\linewidth]{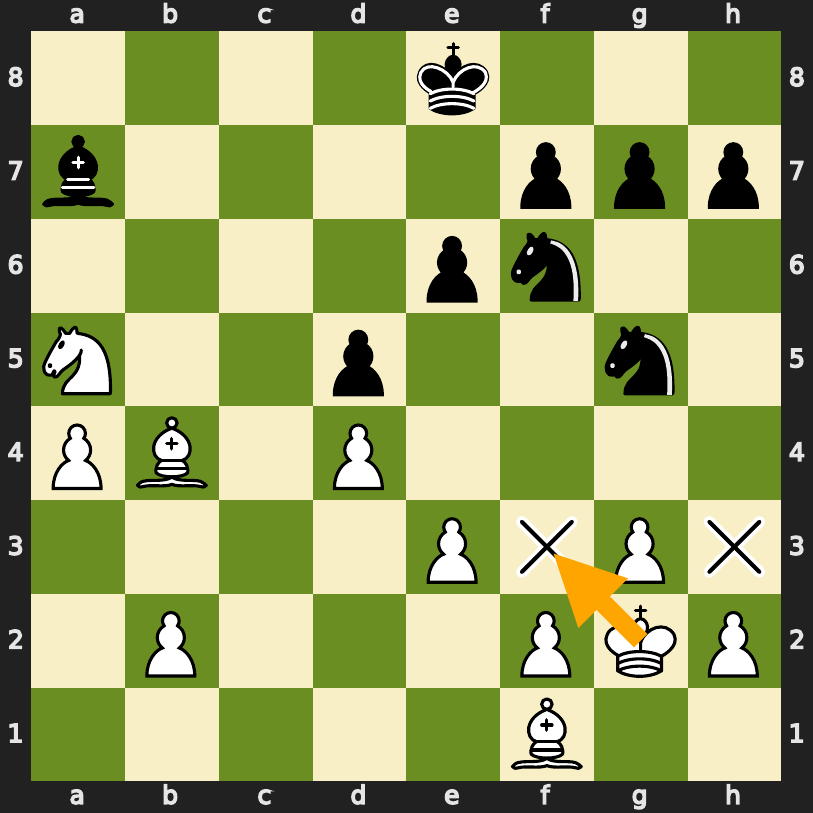}
   \caption{\emph{No Check + King}: The predicted ending square \pos{f3} for the white king is guarded by the black knight at \pos{g5}.} 
   \label{fig:error_no_check_king}
\end{subfigure}
\hspace*{\fill}
\begin{subfigure}{0.43\textwidth}
	\includegraphics[width=\linewidth]{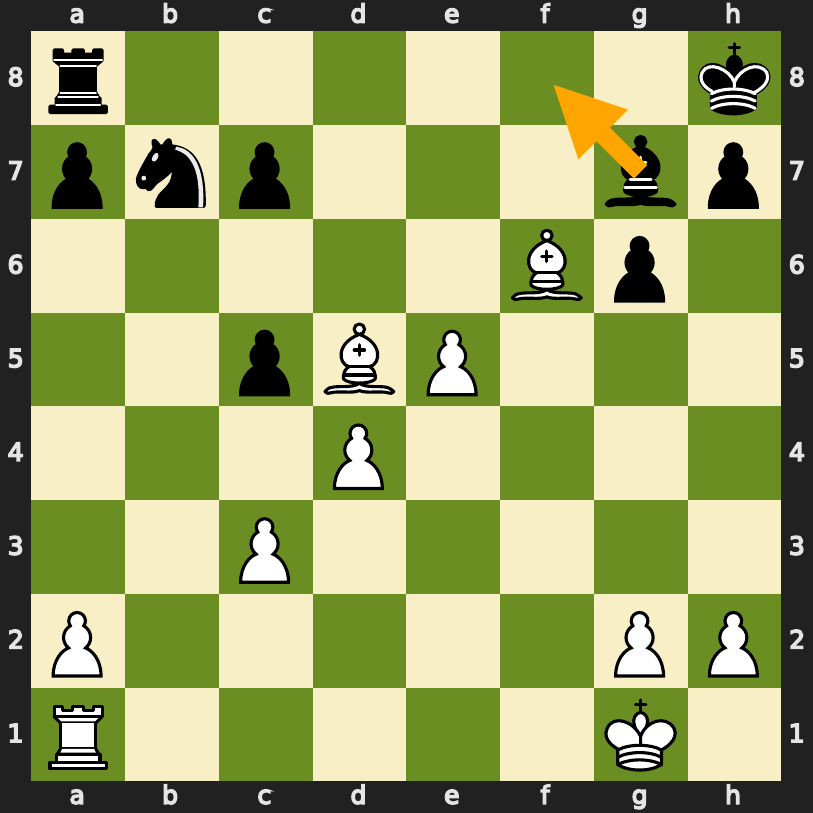}
	\caption{\emph{No Check + Other}: The predicted ending square \pos{f8} for the black bishop exposes its king to the white bishop at \pos{f6}. } 
	\label{fig:error_no_check_no_king}
\end{subfigure}
\hspace*{\fill}
\caption{Four combinations of the king being in check or not, and if the king is moved or not, that can result in Pseudo Legal errors.}
\label{fig:pseudo_legal_errors}
}
\end{figure*}

In this section we conduct a more in-depth analysis of errors made by the UCI model trained with Train-L for the ending square prediction task. We limit our focus to the two main error categories, namely, Pseudo Legal and Path Obstruction.

\begin{table}[t]
	\centering{
		\caption{Pseudo Legal error counts for different categories. For the total column we remove instances with errors of other category.}
		\label{tab:pseudo_legal}
		\begin{tabular}{lcccc}
			\toprule
			Category & \multicolumn{2}{c}{End-Actual} & \multicolumn{2}{c}{End-Other}\\
			& Errors & Total & Errors & Total \\
			\midrule
			Check + King 		& 1	& \phantom{1}27  & \phantom{1}2 & \phantom{1}20	\\
			Check + Other 		& 7	& \phantom{1}26  & 16			& \phantom{1}33 \\ 			 
			No Check + King 	& 4	& 101 			& 31 			& 296	\\ 
			No Check + Other 	& 0 & 835 &	\phantom{1}0  			& 619	\\ 
			\midrule
			Total 				& 12 & 989  	& 49 			& 968\\\bottomrule
		\end{tabular}
	}
	
\end{table}

\begin{table}[ht]
	\caption{Piece type counts for Path Obstruction error category. For the total column we remove instances with errors of other category.}
	\label{tab:path_obs}
	\centering
	\begin{tabular}{lcccc}
		\toprule
		Piece type & \multicolumn{2}{c}{End-Actual} & \multicolumn{2}{c}{End-Other} \\ %
		& Errors & Total & Errors & Total \\
		\midrule
		Rook (\pos{R}) 	& 	3 	& 355	& 17  			& 267    		\\ 
		Knight (\pos{N}) 	& 	1  	& 144	& \phantom{1}1 	& 131 			\\  
		Bishop (\pos{B}) 	& 	1 	& 162	& \phantom{1}3 	& 164 			\\
		Queen (\pos{Q}) 	& 	4  	& 202	& \phantom{1}4  & \phantom{1}99	\\ 
		King (\pos{K}) 	& 	1  	& 124 	& \phantom{1}1  & 284			\\
		\midrule
		Total 		& 	10 	& 987  	& 26 			& 945\\\bottomrule
	\end{tabular}
\end{table}

\subsection{Pseudo Legal Errors}
\label{sec:pseudo_legal}
We conduct our analysis by categorizing instances according to: (a) if the king was in check before the current move, and (b) if the king is being moved in the current move. 
Figure~\ref{fig:pseudo_legal_errors} presents one instance for each of these four categories.
Table~\ref{tab:pseudo_legal} presents the breakdown of errors for the End-Actual and End-Other instances. The key takeaways from the error categorization are: (a) Error counts for ``Check + King" and ``No Check + Other" are relatively low and similar across the two classes of prompts. (b) ``Check + Other" i.e.\ the king is in check and some other piece is moved, has high count for both the splits. The particularly high count for End-Other could be explained by the lack of ``meaningful" moves for certain prompts of this kind. For example, in figure~\ref{fig:error_check_no_king} the prompt asks for the queen at \pos{c8} to move, and the only legal continuation is for the queen to bring itself to the firing line at \pos{c7}. (c) ``No Check + King"  is another common error category. 
The significantly higher error count for End-Other could be due to a combination of the higher frequency of such prompts and the out-of-distribution prompts.

\begin{figure*}[!ht]
	\centering{
		\hspace*{\fill}
		\begin{subfigure}{0.4\textwidth}
			\includegraphics[width=\linewidth]{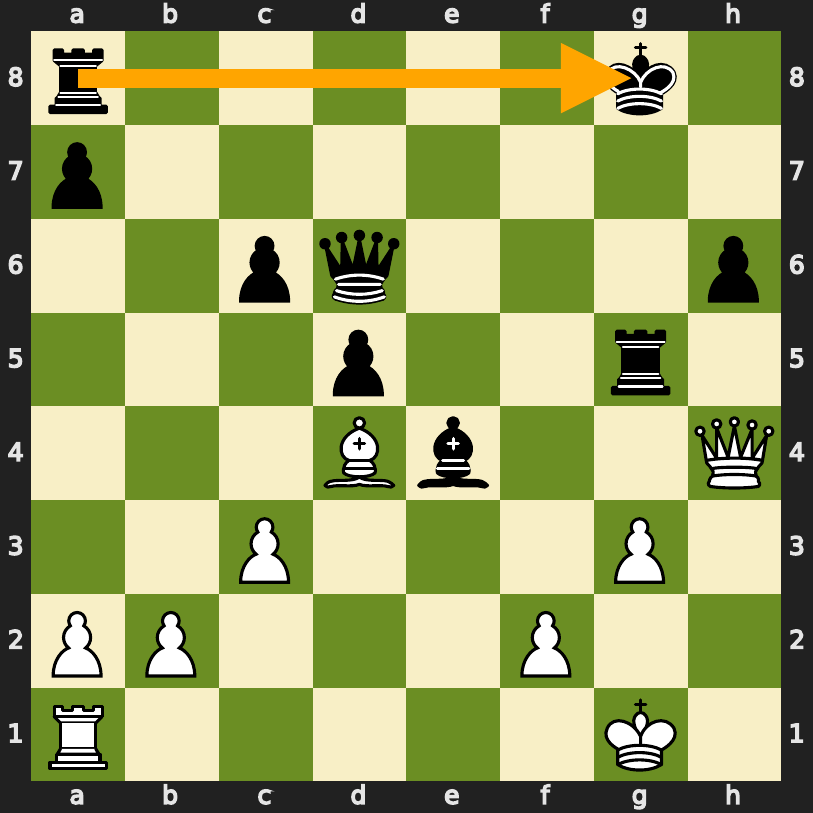}
			\caption{Rook forgets about its own king at \pos{g8}!} 
			\label{fig:rook_path}
		\end{subfigure}
		\hspace*{\fill}
		\begin{subfigure}{0.4\textwidth}
			\includegraphics[width=\linewidth]{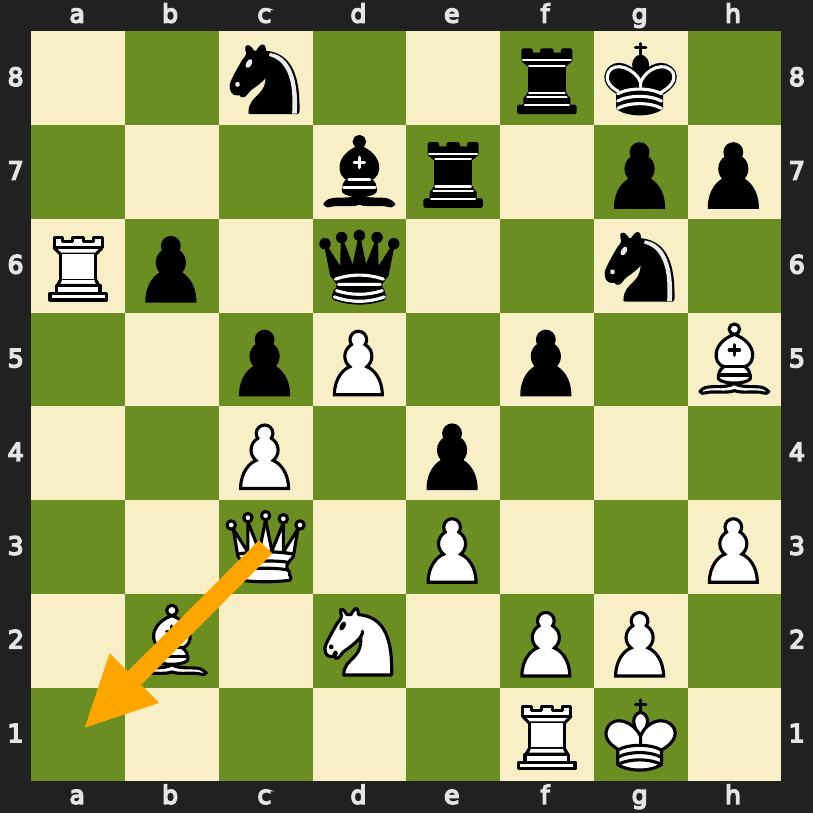}
			\caption{Bishop at \pos{b2} stands in the way of the queen.} 
			\label{fig:queen_path}
		\end{subfigure}
		\hspace*{\fill}
		\\[1em]
		\hspace*{\fill}
		\begin{subfigure}{0.4\textwidth}
			\includegraphics[width=\linewidth]{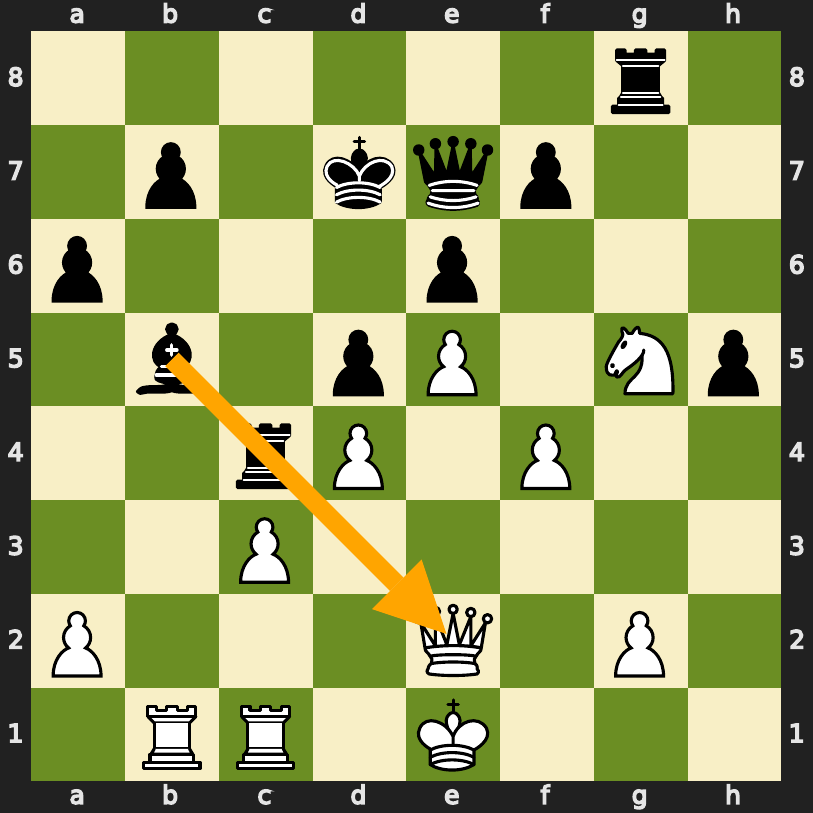}
			\caption{Bishop forgets reality in pursuit of fantasy queen kill!} 
			\label{fig:bishop_path}
		\end{subfigure}
		\hspace*{\fill}
		\begin{subfigure}{0.4\textwidth}
			\includegraphics[width=\linewidth]{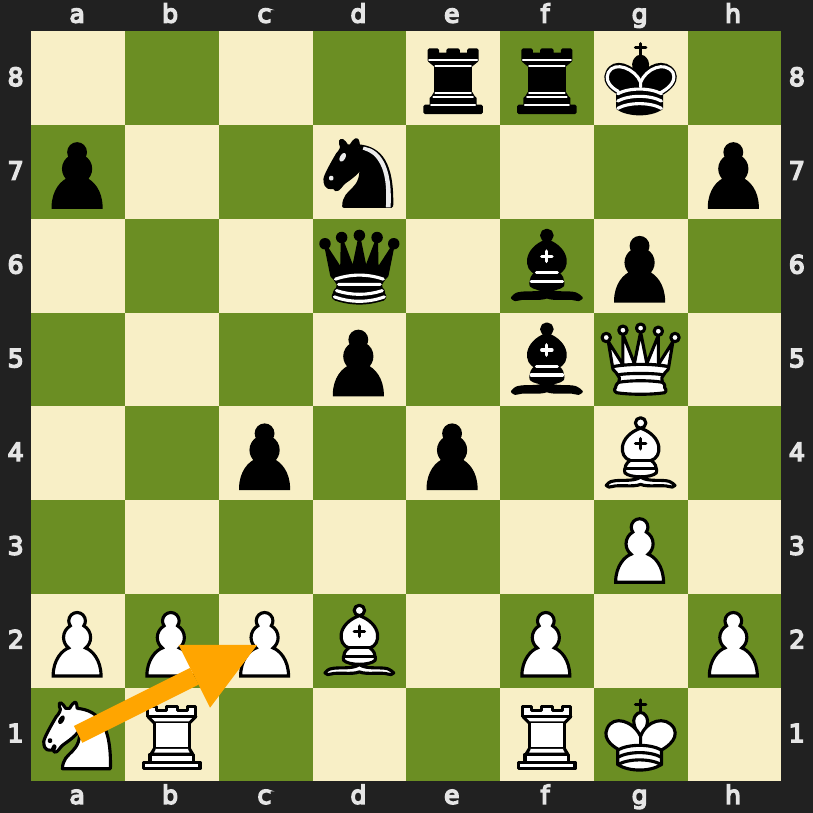}
			\caption{A trapped, frustrated knight is out to kill its own pawn!} 
			\label{fig:knight_path}
		\end{subfigure}
		\hspace*{\fill}
		\caption{Instances of Path Obstruction errors with different piece types.}
		\label{fig:path_obstruction}
	}
\end{figure*}

\begin{figure*}[!ht]
	\centering{
		\hspace*{\fill}
		\begin{subfigure}{0.35\textwidth}
			\includegraphics[width=\linewidth]{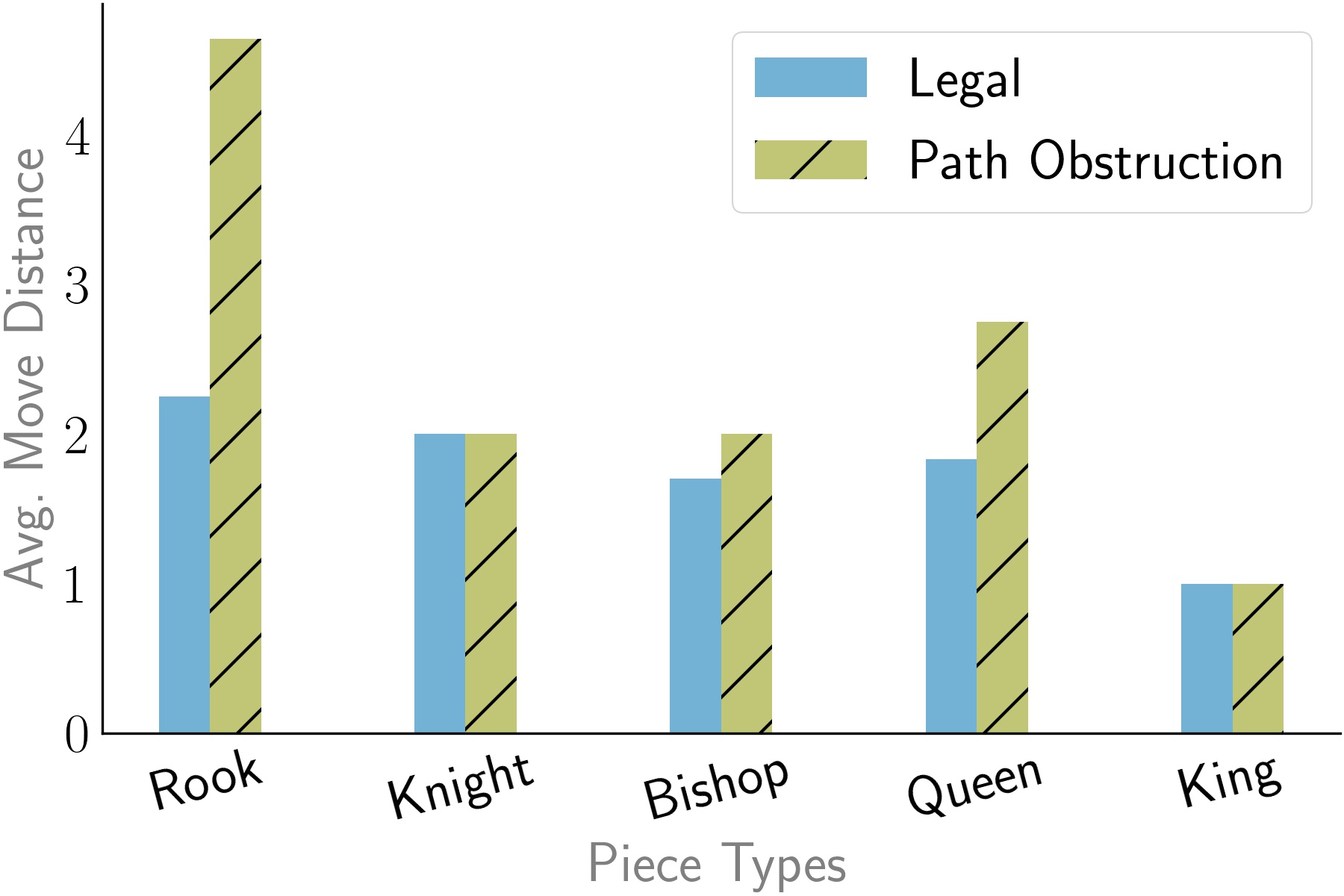}
			\caption{End-Actual} 
			\label{fig:path_length_actual}
		\end{subfigure}
		\hspace*{\fill}
		\begin{subfigure}{0.35\textwidth}
			\includegraphics[width=\linewidth]{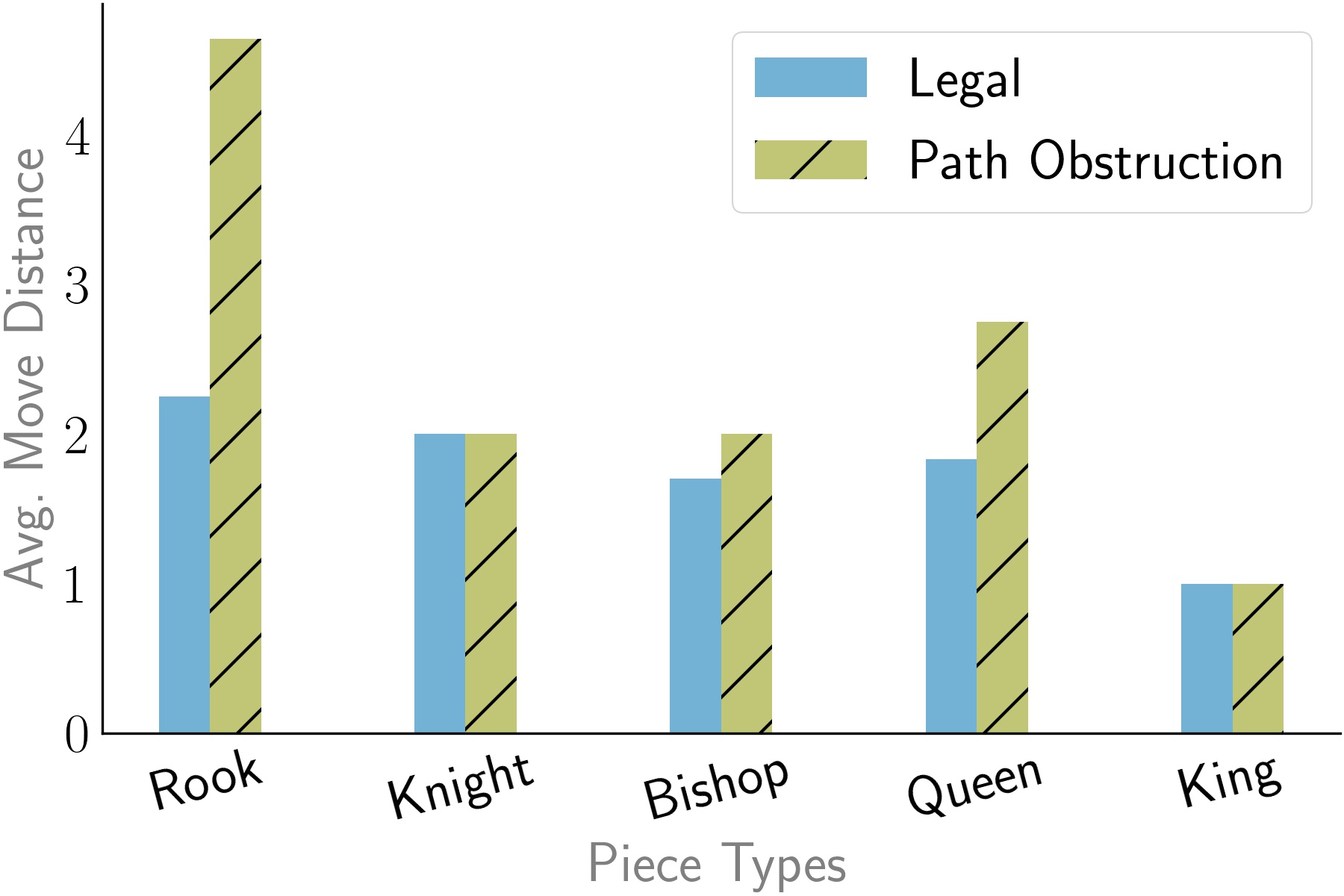}\\
			\caption{End-Other} 
			\label{fig:path_length_other}
		\end{subfigure}
	\hspace*{\fill}
\caption{Comparison of average path length of predicted moves for different piece types when the move is legal vs ones with path obstruction error.}
\label{fig:path_length}
}
\end{figure*}

\subsection{Path Obstruction}
Table~\ref{tab:path_obs} presents the path obstruction errors for different piece types for the End-Actual and End-Other task. Figure~\ref{fig:path_obstruction} represents instances of path obstruction error for different piece types.  
The error counts show that piece types with more dynamic movement except knight i.e. rook, bishop, and queen, have more path obstruction errors  (a knight just needs to avoid landing on its own piece to avoid path obstruction errors). 
These piece types also show a significant increase in frequency of path obstruction errors for End-Other in comparison to End-Actual. As in pseudo legal errors, this could again be due to the out-of-distribution nature of these prompts. Figure~\ref{fig:path_length} shows that the average path length for predicted moves with path obstruction error is significantly higher than the legal predictions for both the kinds of prompts (knight and king have a constant path length). \footnote{Path length is measured in number of king moves i.e. number of moves it will take a king to go from starting to ending square.}

\end{document}